\documentclass[10pt,twocolumn,letterpaper]{article}

\usepackage[pagenumbers]{cvpr}      

\usepackage{amsmath,amssymb,bm}
\usepackage{graphicx}
\usepackage{xspace}
\usepackage[skins]{tcolorbox}
\newtcolorbox{findbox}[1]{enhanced, colback=black!3, colframe=black!3, boxrule=0pt, arc=2pt,
  left=5pt, right=5pt, top=1.5pt, bottom=1.5pt, before skip=3pt, after skip=3pt,
  borderline west={1.8pt}{0pt}{black!75}, fontupper=\small\bfseries,
  before upper={{\normalfont\scriptsize\sffamily\color{black!55}RQ#1}\enspace}}
\newtcolorbox{defbox}{enhanced, colback=accent!5, colframe=accent!55, boxrule=0.5pt,
  arc=1.5pt, left=4pt, right=4pt, top=2pt, bottom=2pt, before skip=6pt, after skip=6pt}
\usepackage{pifont}
\usepackage{booktabs}
\usepackage{colortbl}
\definecolor{oursblue}{HTML}{E3EEFA}
\newcommand{\best}[1]{\textbf{#1}}
\newcommand{\second}[1]{\underline{#1}}
\usepackage{enumitem}
\usepackage{tikz}
\usetikzlibrary{arrows.meta}
\usepackage{caption}
\usepackage{algorithm}
\usepackage{soul}
\usepackage[dvipsnames]{xcolor}
\newcommand{\hlbias}[1]{\textcolor{ForestGreen}{#1}}\usepackage[noend]{algpseudocode}
\usepackage[normalem]{ulem}   
\makeatletter
\g@addto@macro\normalsize{%
  \setlength\abovedisplayskip{4pt plus 1pt minus 2pt}%
  \setlength\belowdisplayskip{4pt plus 1pt minus 2pt}%
  \setlength\abovedisplayshortskip{1pt plus 1pt}%
  \setlength\belowdisplayshortskip{2pt plus 1pt minus 1pt}}
\makeatother

\newcommand{\rqtag}[1]{{\sffamily\bfseries\small\color{MidnightBlue}RQ#1}}
\usepackage{soul}
\definecolor{rqyellow}{HTML}{FFF3B0}
\sethlcolor{rqyellow}
\newcommand{\rqhl}[1]{\textbf{\hl{#1}}}
\definecolor{accent}{HTML}{0F7B6C}
\definecolor{cGen}{HTML}{4C72B0}   
\definecolor{cDes}{HTML}{3A9BB5}   
\definecolor{cDis}{HTML}{B59A2E}   
\definecolor{cVer}{HTML}{DD8452}   
\definecolor{cMea}{HTML}{C44E52}   
\definecolor{cOff}{HTML}{4E9A5F}   
\DeclareRobustCommand{\stepbadgeraw}[2]{\tikz[baseline=-0.62ex]\node[circle,fill=#1,text=white,inner sep=0.7pt,minimum size=1.15em,font=\scriptsize\bfseries]{#2};\,}
\newcommand{\stepbadge}[2]{\texorpdfstring{\stepbadgeraw{#1}{#2}}{}}

\newcommand{\pose}{\textsc{Pose}\xspace}
\newcommand{\eam}{\textsc{Eam}\xspace}
\newcommand{\isc}{ISC\xspace}
\newcommand{\did}{DiD\xspace}

\setlist[itemize]{leftmargin=1.1em, itemsep=2pt, topsep=3pt, parsep=0pt}

\newcommand{\A}{\mathcal{A}}          
\newcommand{\Pset}{\mathcal{P}}       
\newcommand{\Q}{\mathcal{Q}}          
\newcommand{\V}{\mathcal{V}}          
\newcommand{\Sset}{\mathcal{S}}       
\newcommand{\Kset}{\mathcal{K}}       
\newcommand{\vb}{\bm{v}}
\newcommand{\xb}{\bm{x}}
\newcommand{\zb}{\bm{z}}
\newcommand{\ub}{\bm{u}}
\newcommand{\cond}{\bm{c}}
\newcommand{\nul}{\varnothing}
\newcommand{\iscm}{\mathrm{ISC}}
\newcommand{\didm}{\mathrm{DiD}}
\newcommand{\eamm}{\mathrm{EAM}}
\newcommand{\E}{\mathbb{E}}

\usepackage{epigraph}
\usepackage{multirow}

\definecolor{cvprblue}{rgb}{0.21,0.49,0.74}
\usepackage[pagebackref,breaklinks,colorlinks,allcolors=cvprblue]{hyperref}

\def\confName{CVPR}
\def\confYear{2027}

\title{Strike a POSE\\
Auditing and Undoing Gender Depiction Bias in Text-to-Image Generation}

\author{
Rajatsubhra Chakraborty$^{1}$ \quad
Ritabrata Chakraborty$^{2}$ \quad
Xujun Che$^{1}$ \quad
Cori Faklaris$^{1}$ \\
Sijia Qian$^{1}$ \quad
Haopeng Zhang$^{1}$ \quad
Xi Niu$^{1}$ \quad
Depeng Xu$^{1}$ \\[0.5em]
$^{1}$University of North Carolina at Charlotte \qquad
$^{2}$University of Virginia\\
{\tt\small \{rchakra6, xche, cfaklari, sqian, hzhang55, xniu2, dxu7\}@charlotte.edu} \\
{\tt\small zwf5qa@virginia.edu}
}

\begin{document}

\twocolumn[{
\renewcommand\twocolumn[1][]{#1}%
\maketitle
\begin{center}
\captionsetup{type=figure}
\includegraphics[width=\textwidth]{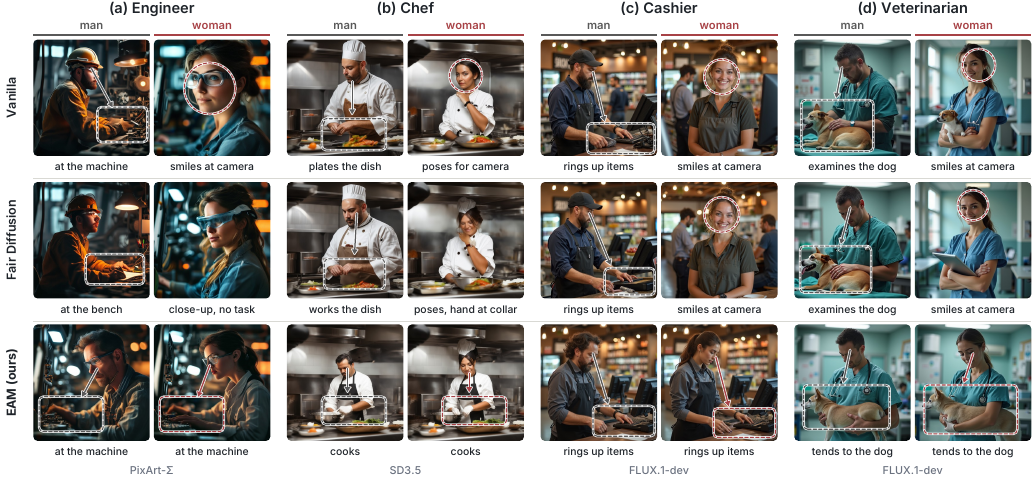}
\captionof{figure}
{\textbf{Gender \hlbias{depiction bias} in Text-to-Image (T2I) models.} We show standard vanilla generation (top row) for the same seed, where each generator depicts the requested gender but the man does the job, while the woman poses for the camera. Fair Diffusion (middle row), even when given the concepts \pose discovers, still leaves the woman posing or idle in every pair. We measure this with an audit \pose and fix these biases with \eam (bottom row). Note how both male and female generations in \eam are depicted similarly.}
\label{fig:teaser}
\end{center}
}]
\begin{abstract}
\vspace{-1.3pt}
Gender bias in text-to-image (T2I) models is usually measured by counting how
many men and women they draw. Yet when a user asks for a woman firefighter, the
model draws one, and there is nothing left to count. What remains is
depiction bias, as men are drawn doing the job while women are drawn posing for
the camera. We introduce \pose, an open-vocabulary audit of how a specified
person is shown. Across various occupations and T2I models, we observe depiction biases consistently. We find
that the layout carrying this bias is committed in the first steps of
sampling, and that a gender followed in those steps tends to stick. Building
on this, we propose Early Attribute Marginalization (\eam), a training-free
sampler that follows the average of the man and woman predictions during the
early steps and then the user's prompt. \eam reduces the structural depiction
gap by 79--97\% while keeping the requested gender in at least 95\% of images.
More broadly, the order in which diffusion models build an image lets us keep
what the user asks for while removing what the stereotype adds.
\end{abstract}
\section{Introduction}
\label{sec:intro}

\providecommand{\hlbias}[1]{\textcolor{accent}{\textbf{#1}}}
\providecommand{\hlprobe}[1]{\textcolor{black!70}{\textbf{#1}}}
\providecommand{\hloffset}[1]{\textcolor{cOff}{\textbf{#1}}}

\epigraph{
    ``Men act and women appear.''
}{
    \footnotesize John Berger, \emph{Ways of Seeing}~\cite{berger1972ways}
}

Text-to-image (T2I) models illustrate news, textbooks, and advertisements on
demand \cite{thomson2025generative,liu2022opal}, and the people depicted in those images shape what viewers learn about who does
what~\cite{aldahoul2025ai,kay2015unequal}. Fairness in these models has mostly been measured by post-hoc counting. Such audits
report the perceived gender, skin tone, or age of the people a neutral prompt
produces~\cite{bianchi2023easily,luccioni2023stable,cho2023dall,naik2023social},
and mitigation methods rebalance that
count~\cite{friedrich2023fair,zhang2023iti,shen2024finetuning,teo2024fairqueue}.
This approach, however, has some limitations. Once a user names the gender of the person, there is nothing left
to count.
Naming the gender is what people do to break a stereotype, say a
woman engineer for a recruitment poster (\cref{fig:teaser}a) or a woman veterinarian for a
textbook (\cref{fig:teaser}d). In standard generation, from the same
seed and the same job, the man works at the machine, plates the dish, rings up
the items, and examines the dog, while the woman smiles or poses for the
camera. We call this \hlbias{depiction bias}, a gap in how a specified person
is shown rather than in who appears (\cref{sec:def}). In classical gender studies and advertising, Berger~\cite{berger1972ways}
and Goffman~\cite{goffman1979gender} establish this pattern, which now persists
in online images~\cite{guilbeault2024online,guilbeault2025age}. This matters
as the perceived fit between a person and a role shapes how competent
that person is judged to be~\cite{heilman1983sex}.

Recent audits show that T2I generators depict women and men
differently~\cite{sun2024smiling,wu2024stable,zhang2023auditing}, but they
measure cues chosen in advance and stop at measurement. Four questions thus arise.
\rqtag{1} \emph{What} differs when no one knows which cues to look for?
\rqtag{2} Is it the \emph{role} or gender alone?
\rqtag{3} \emph{When} in sampling is it decided?
\rqtag{4} Can it be \emph{removed} while the image still shows the requested
person? Unlike rebalancing gender distribution, a remedy here may not change the
person.

We address these questions with \textbf{\pose} (\textbf{P}robing and \textbf{O}ffsetting
\textbf{S}tereotyped \textbf{E}mbodiment). First, we introduce a \hlprobe{probing} framework which generates gender-pair images, extracts gender-blind descriptions, and maps them to a vocabulary of frozen visual concepts. Thereafter, it scores how differently such pairs are depicted.
We observe across models and occupations (\cref{sec:study}) that men are shown doing the work
and women being seen (\rqtag{1}). The posture gap appears only once a role is named and
widens where women are rare in the workforce, and women are drawn younger the
better a job pays (\rqtag{2}). The layout is committed in the first sampling
steps, and a gender followed there tends to stick, so switching the gender word
afterward yields the requested gender in as few as half the images (\rqtag{3}).

To fix this, \pose's \hloffset{offsetting} framework introduces Early Attribute
Marginalization (\eam). For the first $K$ steps it follows the mean of the
\emph{man} and \emph{woman} predictions, sharing the layout without committing
a gender, then the requested prompt alone. It needs no training, weight edits,
or concept lists (\rqtag{4}). Both engineers now work the machine and both
cashiers ring up items (\cref{fig:teaser}, bottom). On all four generators,
\eam reduces structural \isc by 79--97\% and keeps the requested gender in
${\geq}95\%$ of images at about $1.2\times$ the cost.

Our contributions are as follows.
\begin{itemize}
  \item We define \hlbias{depiction bias} for gender-specified requests and
  introduce \textbf{\pose}, an open-vocabulary audit method.
    \item Across multiple models and occupations, we observe that men are drawn
  acting and women appearing, that the posture gap is role-specific and grows
  with occupational scarcity, and that depiction is committed early in
  sampling, where a gender followed early tends to stick.
  \item We introduce \hloffset{\eam}, a training-free architecture-agnostic mitigation method for this \hlbias{depiction bias} that removes most of the structural gap while keeping the requested gender, at about $1.2\times$ the cost.
\end{itemize}
\section{Related Work}
\label{sec:related}

\paragraph{Demographic bias and its mitigation.}
Most fairness work on T2I models asks \emph{who} a prompt produces. Given a
prompt that leaves identity open, audits classify the perceived gender, skin
tone, or age of each generated person and compare the distribution with
parity or labor
statistics~\cite{bianchi2023easily,luccioni2023stable,cho2023dall,naik2023social,seshadri2024bias},
continuing findings for image search~\cite{kay2015unequal} and web-scale
training data~\cite{garcia2023uncurated}, and recently extending to activities,
objects, and contexts~\cite{girrbach2025large}. Mitigation follows the same
framing and rebalances that distribution through prompt
interventions~\cite{bansal2022well}, concept
guidance~\cite{brack2023sega,schramowski2023safe,friedrich2023fair}, learned
tokens~\cite{zhang2023iti}, fine-tuning~\cite{shen2024finetuning}, weight
edits~\cite{gandikota2024unified}, or internal
features~\cite{parihar2024balancing,li2024self,shi2025dissecting,zaleska2026attention}.
Both treat identity as the outcome, so once the prompt names the gender there
is nothing left to count or rebalance. Only Fair Diffusion can be redirected,
since it guides along any concepts a user supplies~\cite{friedrich2023fair},
and we use it as our main baseline.

\paragraph{Depiction in images and generators.}
A named person can still be shown in many ways, and this question predates
generative models. Photographs give men greater facial
prominence~\cite{archer1983faceism}, and captioning and recognition models tie
gender to activities and scene
context~\cite{zhao2017men,hendricks2018women,wang2019balanced}. Five studies
carry it to T2I models. Sun \etal separate representational from
presentational bias and find that DALL$\cdot$E~2 draws women smiling with
their heads tilted down~\cite{sun2024smiling}, a pattern Weinmann \etal confirm
on three commercial platforms~\cite{weinmann2026gender}. Zhang \etal compare a
predefined list of presentation attributes~\cite{zhang2023auditing}, Wu \etal
show with same-seed prompts that the gender word alone changes objects and
layout~\cite{wu2024stable}, and Wan and Chang find that two-person images give
the man the higher-power role, which their FairCritic
corrects~\cite{wan2025male}. Others document sexualized depictions of women in
generators and image-text
models~\cite{ghosh2023person,wolfe2023contrastive}. \Cref{tab:related} shows
what these audits leave open. Two infer gender from the image, which conflates
who is shown with how. All fix their cues in advance, so they cannot say which
postures, gazes, or objects carry the gap in a given occupation, and none
separates what the role adds from gender alone. \pose fixes gender in the
prompt, discovers its concepts from the images, and separates role from
gender with \did.

\paragraph{Open-ended bias discovery.}
Fixing cues in advance is the limitation that open-ended methods address.
OpenBias asks a language model to propose candidate biases and tests them with
visual question answering~\cite{dinca2024openbias}, TIBET derives
prompt-specific bias axes from counterfactual
prompts~\cite{chinchure2024tibet}, and BiasMap localizes hidden demographic
associations through cross-attention~\cite{chakraborty2026biasmap}. All three,
however, inherit the demographic framing and discover \emph{which attributes
vary} across the people a prompt produces. \pose brings open discovery to
depiction, asking \emph{how a fixed person is shown}, and freezes its
vocabulary before any gender comparison.

\paragraph{Timestep-aware intervention.}
Measuring depiction is half the problem. Removing it requires knowing when in
sampling it forms. Diffusion models settle layout in the high-noise steps and
add detail later~\cite{choi2022perception,balaji2022ediff}, which editing
methods exploit by swapping prompts or injecting attention only in chosen
steps~\cite{hertz2023prompt,patashnik2023localizing,liew2022magicmix}, and
guidance by acting in an interval~\cite{kynkaanniemi2024applying}. Fairness
samplers have begun to use this structure, starting from a neutral
prompt~\cite{teo2024fairqueue}, switching the attribute partway
through~\cite{choi2024fair}, or acting on part of the
trajectory~\cite{zaleska2026attention,zhong2026fairt2v}. Each follows a single
prompt early, so the shared layout belongs to the model's default person or to
one gender, and all aim at who appears. \eam follows the average of both
genders instead, sharing the early layout without committing to either, and
is the only mitigation in \cref{tab:related} that targets depiction for a
single named person.

\begin{table}[t]
\centering
\footnotesize
\setlength{\tabcolsep}{2.4pt}
\caption{\textbf{Audits of depiction in text-to-image models.} Whether gender
is requested in the prompt or inferred from the image, whether the concepts
are discovered from the images rather than fixed in advance, which cues are
measured, and whether a mitigation targets depiction.}
\label{tab:related}
\begin{tabular}{lcccc}
\toprule
Audits & Gender & Open & Cues & Mitigates \\
 & requested & discovery & measured & depiction \\
\midrule
Sun \etal~\cite{sun2024smiling}          & \ding{55} & \ding{55} & facial & \ding{55} \\
Weinmann \etal~\cite{weinmann2026gender} & \ding{55} & \ding{55} & facial & \ding{55} \\
Zhang \etal~\cite{zhang2023auditing}     & \ding{51} & \ding{55} & attire & \ding{55} \\
Wu \etal~\cite{wu2024stable}             & \ding{51} & \ding{55} & layout & \ding{55} \\
Wan and Chang~\cite{wan2025male}         & \ding{51} & \ding{55} & roles  & \ding{51} \\
\midrule
\rowcolor{oursblue}
\textbf{\pose (ours)} & \ding{51} & \ding{51} & \textbf{\begin{tabular}[c]{@{}c@{}}pose, gaze,\\ activity, attire\end{tabular}} & \ding{51} \\
\bottomrule
\end{tabular}
\end{table}

\section{Method}
\label{sec:pose}

\providecommand{\hlprobe}[1]{\textcolor{black!70}{\textbf{#1}}}
\providecommand{\hloffset}[1]{\textcolor{cOff}{\textbf{#1}}}
\newcommand{\algchange}[1]{\colorbox{cOff!14}{$\displaystyle #1$}}
\newcommand{\algchangetext}[1]{\colorbox{cOff!14}{#1}}
\pose has two halves (\cref{fig:pipeline}). \hlprobe{Probing} discovers and
measures how a generator shows a person whose gender the user has specified
(\cref{sec:probe}), and \hloffset{offsetting} removes the part of that
depiction the request did not ask for (\cref{sec:eam}).

\begin{figure*}[t]
\centering
\includegraphics[width=\textwidth]{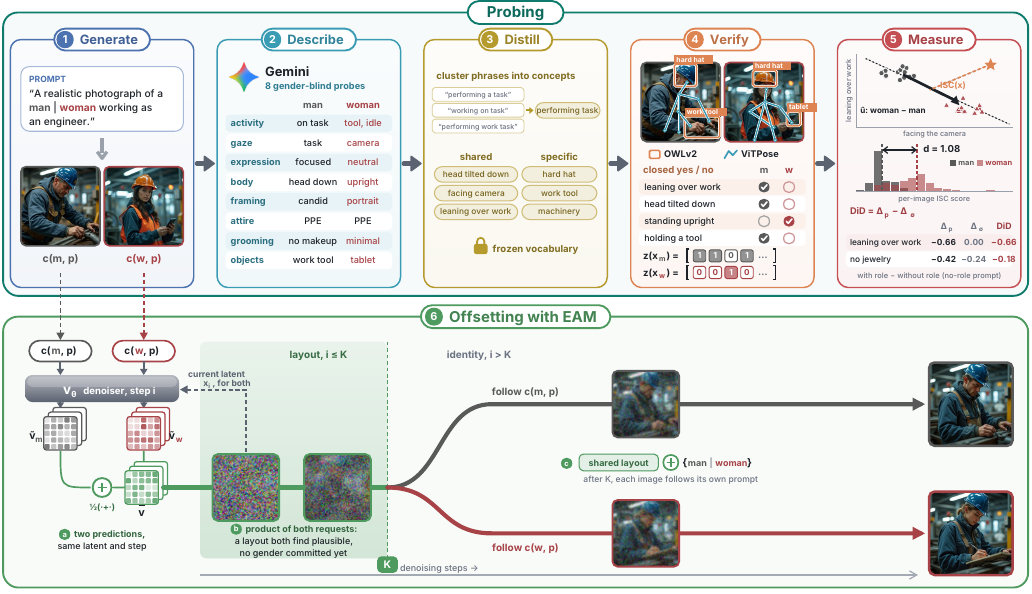}
\vspace{-1.4em}
\caption{\textbf{Overview of \pose.} We show how \pose finds and removes
\hlbias{depiction bias}. In \hlprobe{probing} (top), we \stepbadge{cGen}{1}~generate a
same-seed \emph{man} and \emph{woman}, \stepbadge{cDes}{2}~describe them with
gender-blind probes, \stepbadge{cDis}{3}~distill a frozen vocabulary,
\stepbadge{cVer}{4}~verify each concept, and \stepbadge{cMea}{5}~score each
image against the stereotype direction. In \hloffset{offsetting} (bottom),
\stepbadge{cOff}{6}~\eam follows the average of both predictions for the first
$K$ steps, then the requested prompt. Best viewed in zoom.}
\label{fig:pipeline}
\vspace{-0.6em}
\end{figure*}

\subsection{Preliminaries}
\label{sec:prelim}

\paragraph{\stepbadge{cGen}{1}Paired requests.}
A generator $G$ maps a prompt $\cond$ and a seed $s$ to an image
$\xb = G(\cond, s)$. With attributes $\A = \{\text{m}, \text{w}\}$ and
occupations $\Pset$, a request is
\par\vspace{1pt}{\centering\small $\cond(a, p) =$ ``A realistic photograph of a $\langle a \rangle$ working as $\langle p \rangle$.''\par}\vspace{1pt}\noindent
and we also use the reference prompt $\cond(a, \nul)$, which names no
occupation, and the neutral prompt $\cond(\text{person}, p)$, which names no
gender. The pair $\xb^{\text{m}}_{p,s} = G(\cond(\text{m},p), s)$ and
$\xb^{\text{w}}_{p,s} = G(\cond(\text{w},p), s)$ starts from the same noise
$\xb_1 = \bm{\epsilon}_s$, so the gender word is the only cause of a
difference within it. A sampler takes $N$ steps over noise levels
$\sigma_1 > \dots > \sigma_N$. At step $i$ the denoiser predicts
$\vb_\theta(\xb_i, \sigma_i, \cond)$, and classifier-free
guidance~\cite{ho2022classifier} of scale $\lambda$ gives
\begin{equation}
\tilde{\vb}(\xb, \sigma, \cond) = \vb_\theta(\xb, \sigma, \nul)
+ \lambda \big[ \vb_\theta(\xb, \sigma, \cond) - \vb_\theta(\xb, \sigma, \nul) \big],
\label{eq:cfg}
\end{equation}
followed by the solver update $\xb_{i+1} = \Phi_i\big(\xb_i, \tilde{\vb}(\xb_i, \sigma_i, \cond)\big)$.
A guidance-distilled model computes $\tilde{\vb}$ in one pass.

\subsection{Depiction bias}
\label{sec:def}

\begin{defbox}
\textbf{Definition} (Depiction bias). Let $\bm{\phi}(\xb)$ be a vector of
visual facts about how the person in $\xb$ is shown that the
request does not determine, such as pose, gaze, framing, held objects,
expression, and attire. The depiction bias of $G$ for occupation $p$ is the
expected difference within pairs,
\begin{equation*}
\bm{\beta}_p = \E_{s}\big[\, \bm{\phi}(\xb^{\text{w}}_{p,s}) - \bm{\phi}(\xb^{\text{m}}_{p,s}) \,\big],
\end{equation*}
and $G$ has depiction bias when $\bm{\beta}_p \neq \bm{0}$.
\end{defbox}

Since gender is fixed by the request, $\bm{\beta}_p$ can be large for a
generator that always draws the person it was asked for. Cues that signal
gender itself, such as hair and facial hair, are the request being honored and
are excluded from $\bm{\phi}$. With $\bm{\beta}_{\nul}$ the same quantity under
the reference prompts,
\begin{equation}
\bm{\beta}_p = \underbrace{\bm{\beta}_{\nul}}_{\text{gender-only}}
+ \underbrace{\big(\bm{\beta}_p - \bm{\beta}_{\nul}\big)}_{\text{role-specific}} ,
\label{eq:decomp}
\end{equation}
separating how the generator draws a woman in general from what the
occupation adds.

\subsection{Probing}
\label{sec:probe}
\label{sec:measure}

Probing estimates $\bm{\phi}$ without a list of cues written in advance, and
measures $\bm{\beta}_p$ and its two terms in \cref{eq:decomp} with two
metrics.

\paragraph{\stepbadge{cDes}{2}Gender-blind description.}
A vision-language model $D$~\cite{gemini2025} answers eight probes
$\Q = \{$activity, gaze, expression, body, framing, attire, grooming,
objects$\}$, returning free text $t_q(\xb)$ for each $q \in \Q$. It is told to
describe only what is visible, to call the subject \emph{the person}, and never
to infer gender, age, race, attractiveness, personality, competence, warmth, or
status, so that a difference in the text reflects a difference in the pixels.

\paragraph{\stepbadge{cDis}{3}A frozen two-tier vocabulary.}
Phrases from the descriptions $t_q(\xb)$ of a discovery seed set
$\Sset_{\text{disc}}$ are embedded with a sentence encoder and clustered within
each occupation and probe at cosine similarity $0.85$, keeping negations
distinct, as in \emph{holding a tool} and \emph{not holding a tool}. With
$f_p(k)$ the fraction of images of occupation $p$ whose descriptions contain
concept $k$,
\begin{equation}
\begin{aligned}
\V_{\text{sh}} &= \big\{\, k \;\big|\; |\{ p \in \Pset \mid f_p(k) \ge \tau \}| \ge 3 \,\big\}, \\
\V_p &= \big\{\, k \notin \V_{\text{sh}} \;\big|\; f_p(k) \ge \tau \,\big\},
\end{aligned}
\label{eq:vocab}
\end{equation}
are the concepts shared across occupations and specific to one. Near-duplicate
concepts (token Jaccard $\ge 0.6$) are merged, and the vocabulary is frozen
before any gender comparison, so the concepts we test are not chosen by the
gaps they produce.

\paragraph{\stepbadge{cVer}{4}Verification and layers.}
Because an open description can omit what is present, $D$ answers a closed
yes or no question for every concept in $\V_{\text{sh}} \cup \V_p$, giving
$z_k(\xb) \in \{0,1\}$. Measurements independent of $D$ are added. An
open-vocabulary detector~\cite{minderer2023scaling} locates the objects named
under the objects probe, an object counts as held when its box contains a wrist
keypoint from a pose estimator~\cite{xu2022vitpose}, head pitch and torso lean
come from the keypoints, and perceived age is estimated separately. The result
is $\zb(\xb) \in \mathbb{R}^{M}$, our estimate of $\bm{\phi}(\xb)$ together
with the marking cues. Each coordinate inherits the probe $q \in \Q$ it came
from, which assigns it to one of three layers
$L \in \{\text{str}, \text{pres}, \text{mark}\}$ fixed before any analysis.
\emph{Structure} (activity, gaze, body, framing, objects, pose) and
\emph{presentation} (expression, attire, non-gendered grooming such as
jewelry) instantiate $\bm{\phi}$. \emph{Marking} (hair length, facial hair,
cosmetics) is the request and is tracked only to confirm that a mitigation
keeps it.

\paragraph{\stepbadge{cMea}{5}Role-specific gap (\did).}
The paired gap for occupation $p$ and concept $k$ over a seed set $\Sset$,
\begin{equation}
\Delta_{p,k} = \frac{1}{|\Sset|} \sum_{s \in \Sset}
\Big[ z_k\big(\xb^{\text{w}}_{p,s}\big) - z_k\big(\xb^{\text{m}}_{p,s}\big) \Big],
\label{eq:gap}
\end{equation}
estimates coordinate $k$ of $\bm{\beta}_p$, and the difference-in-differences
$\didm_{p,k} = \Delta_{p,k} - \Delta_{\nul,k}$ against the reference prompts
estimates the role-specific term of \cref{eq:decomp}. A concept is
role-specific in $p$ when the 95\% bootstrap interval of $\didm_{p,k}$ over
seeds excludes zero. To score a mitigation we report the role-specific gap it
leaves,
$\sum_k |\didm^{\text{method}}_{p,k}| / \sum_k |\didm^{\text{vanilla}}_{p,k}|$
over structural concepts. The magnitude matters. A method that reverses a gap,
drawing the man posing instead of the woman, has not removed it, and the ratio
then exceeds 100\%.

\paragraph{Individual stereotype congruence (\isc).}
\isc summarizes all concepts in one score per image. For occupation $p$,
features are standardized with the mean $\bm{\mu}_{-p}$ and standard deviation
$\bm{\omega}_{-p}$ of the images in $\Sset_{\text{disc}}$ of all \emph{other}
occupations,
$\tilde{\zb}(\xb) = (\zb(\xb) - \bm{\mu}_{-p}) \oslash \bm{\omega}_{-p}$, so no
occupation defines its own stereotype. With
$\bm{\delta}_{-p} = \bar{\tilde{\zb}}^{\,\text{w}}_{-p} - \bar{\tilde{\zb}}^{\,\text{m}}_{-p}$
the woman$-$man difference of those means,
\begin{equation}
\ub_{-p} = \bm{\delta}_{-p} / \|\bm{\delta}_{-p}\|_2,
\qquad
\iscm(\xb) = \big\langle \tilde{\zb}(\xb),\, \ub_{-p} \big\rangle ,
\label{eq:isc}
\end{equation}
and restricting both to the coordinates of layer $L$ gives $\iscm_L$. A
positive score means the image is depicted the way the generator tends to
depict women. For each pair,
$g_{p,s} = \iscm(\xb^{\text{w}}_{p,s}) - \iscm(\xb^{\text{m}}_{p,s})$. Since
\isc is linear, $\bar{g}_p$ is the standardized estimate of $\bm{\beta}_p$
projected onto $\ub_{-p}$, so the effect size $d_p = \bar{g}_p / \omega_g$,
with $\omega_g$ the standard deviation of $g$ over vanilla pairs held fixed
across methods, summarizes depiction bias in one number, and $d_p = 0$ means
the pair is depicted alike. We report $d_p$ on the structure layer,
$\iscm_{\text{str}}$, unless stated otherwise.

\subsection{Offsetting}
\label{sec:eam}

Diffusion models settle layout in the high-noise
steps~\cite{choi2022perception,balaji2022ediff}, and \cref{sec:study_early}
shows that the depiction is committed there too, together with any gender the
early steps follow. Offsetting aims to drive the structural coordinates of
$\bm{\beta}_p$ to zero while keeping the marking the user asked for. Early
Attribute Marginalization (\eam) therefore follows the average of both genders
during the early steps and the requested gender afterwards.

\paragraph{\stepbadge{cOff}{6}The sampler.}
For a request $(a^\star, p)$ and a distribution $\bm{\pi}$ over $\A$, uniform
unless stated otherwise,
\begin{equation}
\vb^{\eamm}_i =
\begin{cases}
\sum_{a \in \A} \pi_a\, \tilde{\vb}\big(\xb_i, \sigma_i, \cond(a, p)\big) & i \le K, \\[2pt]
\tilde{\vb}\big(\xb_i, \sigma_i, \cond(a^\star, p)\big) & i > K ,
\end{cases}
\label{eq:eam}
\end{equation}
with switch step $K$ the only hyperparameter, so $K = 0$ is the vanilla
sampler. \Cref{alg:eam} highlights the only change, lines~\ref{alg:if}
and~\ref{alg:avg}. Since \cref{eq:cfg} is affine in the conditional prediction
and $\sum_a \pi_a = 1$, the average shares the unconditional pass of
line~\ref{alg:uncond} and is guided once in line~\ref{alg:update}. For a
guidance-distilled model, line~\ref{alg:uncond} is dropped and
line~\ref{alg:update} uses $\vb^{\text{c}}_i$ directly.

\begin{algorithm}[t]
\caption{Early Attribute Marginalization (\eam). \algchangetext{Highlighted steps}
denote changes to vanilla sampling.}
\label{alg:eam}
\small
\begin{algorithmic}[1]
\Require request $(a^\star, p)$, seed $s$, switch step $K$, weights $\bm{\pi}$ over $\A$
\State $\xb_1 \gets \bm{\epsilon}_s$
\For{$i = 1, \dots, N$}
  \State $\vb^{\nul}_i \gets \vb_\theta(\xb_i, \sigma_i, \nul)$ \Comment{unconditional pass}\label{alg:uncond}
  \If{\algchange{i \le K}} \Comment{layout phase}\label{alg:if}
    \State \algchange{\vb^{\text{c}}_i \gets \textstyle\sum_{a \in \A} \pi_a\, \vb_\theta\big(\xb_i, \sigma_i, \cond(a,p)\big)}\label{alg:avg}
  \Else \Comment{identity phase}\label{alg:else}
    \State $\vb^{\text{c}}_i \gets \vb_\theta\big(\xb_i, \sigma_i, \cond(a^\star,p)\big)$\label{alg:req}
  \EndIf
  \State $\xb_{i+1} \gets \Phi_i\big(\xb_i,\ \vb^{\nul}_i + \lambda(\vb^{\text{c}}_i - \vb^{\nul}_i)\big)$\label{alg:update}
\EndFor
\State \Return $\mathrm{Decode}(\xb_{N+1})$
\end{algorithmic}
\end{algorithm}

\paragraph{What the average does.}
At a fixed noise level each prediction is affine in the score
$\nabla \log \rho_\sigma(\xb \,|\, \cond)$~\cite{song2021score}, with
$\rho_\sigma$ the noised data density. With $\cond_a = \cond(a,p)$, the
unguided early prediction of \cref{eq:eam} therefore follows
\begin{equation}
\textstyle\sum_{a} \pi_a \nabla \log \rho_\sigma(\xb \,|\, \cond_a)
= \nabla \log \prod_{a} \rho_\sigma(\xb \,|\, \cond_a)^{\pi_a} ,
\label{eq:poe}
\end{equation}
a product of the conditionals, the composition composable diffusion uses for
a conjunction of concepts~\cite{liu2022compositional}. A product favors layouts
that every request finds plausible and commits to no single gender, whereas
following one gender commits each sample to that gender's layout and, as
\cref{sec:study_early} shows, often to the gender itself. The identity holds
under guidance with the unconditional density raised to $1-\lambda$.

\paragraph{Choosing the switch step.}
A larger $K$ shares more of the trajectory but can lose the requested gender.
With $F(K)$ the share of images on a calibration seed set $\Sset_{\text{cal}}$
in which zero-shot CLIP~\cite{radford2021learning} recognizes the requested
gender, we take the largest of the candidates $\Kset$ within $\delta$ of the
vanilla sampler,
\begin{equation}
K^\star = \max \big\{\, K \in \Kset \;\big|\; F(K) \ge F(0) - \delta \,\big\},
\label{eq:kstar}
\end{equation}
one per generator, and give an occupation its own $K$ only if its fidelity at
$K^\star$ falls more than $\delta'$ below vanilla.

\paragraph{Relation to other samplers.}
Prompt queuing~\cite{teo2024fairqueue} and attribute
switching~\cite{choi2024fair} also replace only line~\ref{alg:avg}, with the
neutral prompt $\cond(\text{person},p)$ or one fixed gender $\cond(a',p)$, so
their early layout belongs to the model's default person or to one gender,
while that of \eam belongs to neither. \Cref{eq:eam} extends directly to
$|\A| > 2$ and to intersectional attributes with a product distribution
$\bm{\pi}$.
\section{Experiments}
\label{sec:exp}
\label{sec:study}

We describe the setup, answer the four questions of \cref{sec:intro} with one
finding each, and close with what they imply. Implementation details are in
Supp.~S1 and~S2.

\providecommand{\na}{\textcolor{black!35}{n/a}}

\begin{figure*}[t]
\centering
\includegraphics[width=\linewidth]{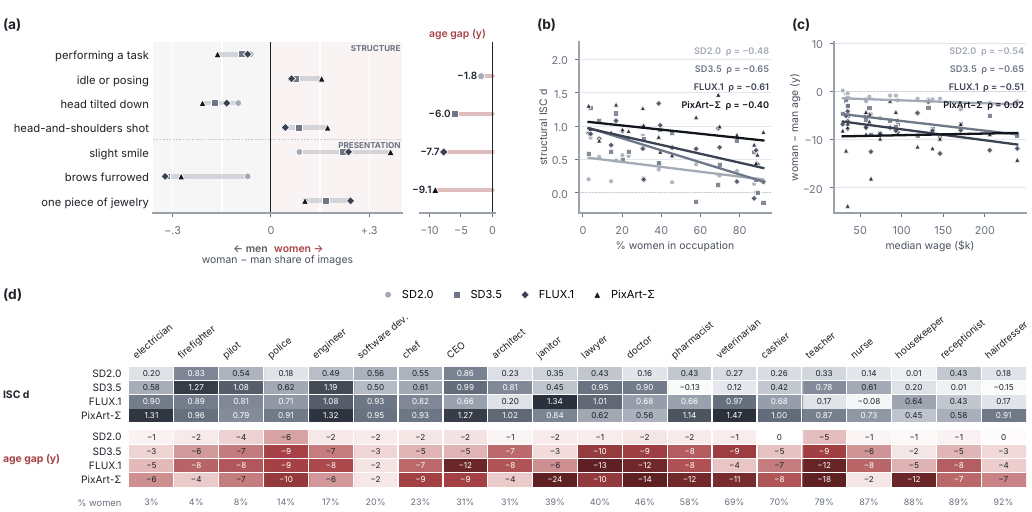}
\caption{\textbf{Findings of \pose.} (a)~Pooled woman$-$man gaps from the same
closed questions in every generator (left) and the mean age gap (right).
(b)~Structural \isc against the share of women. (c)~Age gap against median
wage. For both (b,~c)~we have one point per occupation, with least-squares fits and Spearman's
$\rho$. (d)~Structural \isc and age gap per occupation, sorted by the share of
women.}
\label{fig:findings}
\end{figure*}

\begin{table*}[t]
\centering
\caption{\textbf{Quantitative results for depiction bias.} Cost refers to network function evaluations (NFEs) relative to vanilla implementation. Best results are shown in \textbf{bold} and second-best results are \underline{underlined}. $^{*}$Rewrites the prompt.
}
\label{tab:main}
\vspace{-0.6em}
\small
\setlength{\tabcolsep}{3.2pt}
\renewcommand{\arraystretch}{1.05}
\begin{tabular}{l ccc ccc ccc ccc c}
\toprule
 & \multicolumn{3}{c}{SD2.0} & \multicolumn{3}{c}{SD3.5} & \multicolumn{3}{c}{FLUX.1} & \multicolumn{3}{c}{PixArt-$\Sigma$} & \\
\cmidrule(lr){2-4}\cmidrule(lr){5-7}\cmidrule(lr){8-10}\cmidrule(lr){11-13}
Method & ISC$\downarrow$ & DiD$\downarrow$ & GF$\uparrow$ & ISC$\downarrow$ & DiD$\downarrow$ & GF$\uparrow$ & ISC$\downarrow$ & DiD$\downarrow$ & GF$\uparrow$ & ISC$\downarrow$ & DiD$\downarrow$ & GF$\uparrow$ & Cost$\downarrow$ \\
\midrule
Vanilla & 0.42 & 100 & 97.4 & 0.61 & 100 & 100.0 & 0.56 & 100 & 100.0 & 0.83 & 100 & 100.0 & 1.0 \\
\midrule
Metric noise floor & 0.00 & -- & -- & 0.00 & -- & -- & 0.00 & -- & -- & 0.00 & -- & -- & -- \\
Edit oracle (upper bound) & 0.01 & 3 & 96.8 & 0.01 & 2 & 99.4 & 0.02 & 4 & 99.2 & 0.02 & 5 & 99.0 & -- \\
\midrule
Prompt intervention$^{*}$~\cite{bansal2022well} & 0.35 & 84 & \best{97.0} & 0.49 & 82 & \second{99.5} & 0.45 & 86 & \best{99.7} & 0.68 & 85 & \second{99.6} & \best{1.0} \\
Prompt queuing~\cite{teo2024fairqueue} & 0.20 & 67 & 88.6 & 0.25 & 61 & 91.2 & 0.22 & 65 & 89.4 & 0.35 & 69 & 90.5 & \best{1.0} \\
Attribute switching~\cite{choi2024fair} & 0.16 & \second{59} & 78.4 & \second{0.21} & \second{55} & 82.1 & \second{0.15} & \second{58} & 76.3 & \second{0.28} & \second{63} & 79.8 & \best{1.0} \\
BiasMap~\cite{chakraborty2026biasmap} & 0.37 & 91 & 92.8 & \na & \na & \na & \na & \na & \na & \na & \na & \na & -- \\
Fair Diffusion~\cite{friedrich2023fair} & \best{0.02} & 129 & 95.0 & 0.28 & 71 & 97.5 & 0.31 & 105 & \second{99.1} & 0.39 & 88 & 98.4 & 3--5 \\
\midrule
\rowcolor{oursblue} \eam (ours) & \second{0.03} & \best{38} & \second{95.2} & \best{0.02} & \best{28} & \best{99.9} & \best{0.12} & \best{42} & 98.6 & \best{0.14} & \best{40} & \best{100.0} & \second{1.2} \\
\bottomrule
\end{tabular}
\vspace{-0.8em}
\end{table*}

\begin{figure*}[t]
\centering
\includegraphics[width=0.97\linewidth]{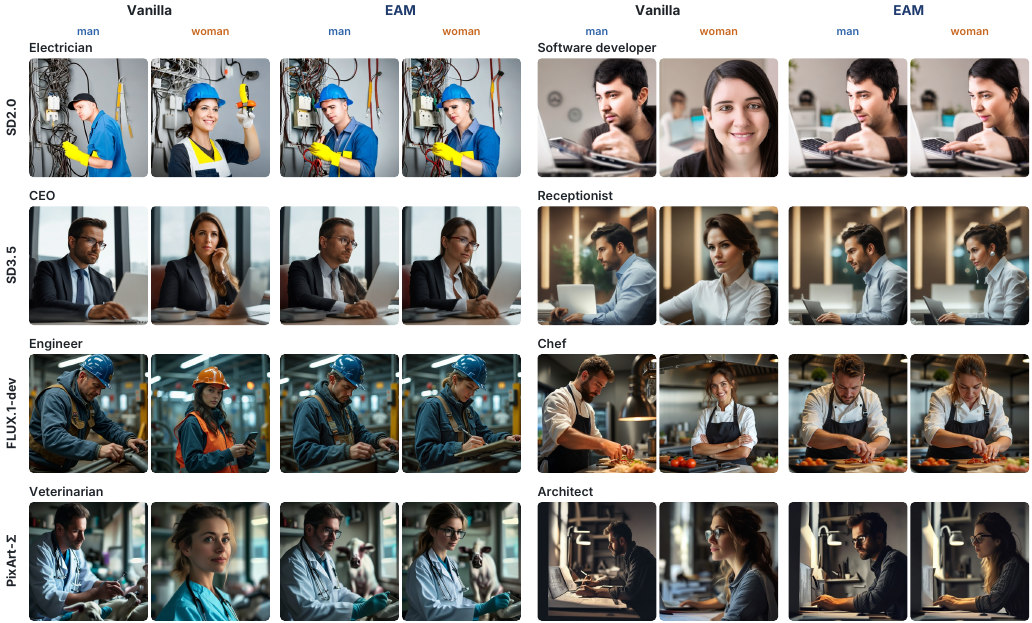}
\caption{\textbf{Qualitative results for \hlbias{depiction bias}.} In each case for vanilla generation (left) the man works and
the woman poses or smiles at the camera, even as a receptionist or
veterinarian. With \eam (right), both share one working layout, and each keeps the requested gender.}
\label{fig:qual_eam}
\end{figure*}

\begin{figure}[t]
\centering
\includegraphics[width=\linewidth]{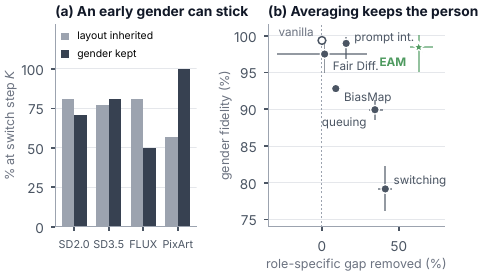}
\caption{\textbf{Impact of what the early steps follow.} (a)~When the man's
prompt drives the first $K$ steps, most of the layout is inherited and the
requested woman often fails to appear (\rqtag{3}). (b)~Gap removed against
gender fidelity, averaged over generators (\cref{tab:main}). Only \eam removes
most of the gap and keeps the person (\rqtag{4}).}
\label{fig:rq34}
\end{figure}

\subsection{Setup}
\label{sec:study_setup}

\paragraph{Generators and prompts.}
We study four generators from three denoiser families, SD2.0 with a
UNet~\cite{rombach2022high}, PixArt-$\Sigma$ with a cross-attention
transformer~\cite{chen2024pixart}, and SD3.5 and FLUX.1-dev with MM-DiT
backbones~\cite{esser2024scaling,flux2024}, each with its default sampler and
guidance. The 20 occupations span U.S. workforce composition from electricians
at 2.9\% women to hairdressers at 92.1\%~\cite{bls2023cps}. Each occupation has
50 discovery seed pairs, each gender 50 reference seeds, and mitigation is
scored on 50 further held-out pairs per occupation. \eam uses $K^\star$ = 20 of
50 steps on SD2.0, 10 of 28 on SD3.5, 6 of 28 on FLUX.1-dev, and 8 of 20 on
PixArt-$\Sigma$, chosen on separate calibration seeds with $\delta = 2$ and
$\delta' = 10$ (\cref{eq:kstar}).

\paragraph{References, baselines, and metrics.}
The metric noise floor scores two independent images of the same gender as a
pair, the value sampling noise alone produces, and the edit oracle edits the
man into a woman so that the pair shares the man's layout by construction.
Prompt intervention appends an instruction to the request~\cite{bansal2022well}.
Prompt queuing~\cite{teo2024fairqueue} and attribute
switching~\cite{choi2024fair} follow the neutral prompt or one fixed gender for
the first $K^\star$ steps, so they differ from \eam only in line~\ref{alg:avg}.
BiasMap~\cite{chakraborty2026biasmap} needs UNet cross-attention and runs on
SD2.0 only, and Fair Diffusion~\cite{friedrich2023fair} is given the concepts
\pose discovers, information \eam never sees (Supp.~S6). We report structural
\isc $d$, the role-specific \did left, and gender fidelity (GF), the share of
images in which zero-shot CLIP~\cite{radford2021learning} recognizes the
requested gender. Supp.~S3 checks the describer against the instruments.

\subsection{RQ1. What differs when cues are unknown?}
\label{sec:study_asym}
\label{sec:rq1}

\rqhl{An open vocabulary finds the gap in posture, activity, and held objects
as well as in faces, and every generator shares it.} In all four generators,
women are less often shown performing a task or looking down at the work and
more often idle, posing, or framed as a portrait (\cref{fig:findings}a). The
discovered concepts move beyond facial cues. Men receive
the tools of the role, the hose, the mop, the gavel, the controls, and also
watches and glasses where nothing in the prompt calls for them. Facial cues
differ too, as women smile more and furrow their brows less. No pooled gap
changes sign across generators, and the direction rarely flips even where
women dominate the workforce, so that PixArt-$\Sigma$ draws the man doing the
work in 18 of 20 occupations. The generators agree on the script but not on
where it applies most, with a rank correlation of per-occupation \isc of only
0.04 to 0.45 (\cref{fig:findings}d), so an audit of one generator does not
transfer to another.

\subsection{RQ2. Is it the role or gender alone?}
\label{sec:study_role}
\label{sec:study_fit}
\label{sec:rq2}
\begin{figure}[t]
\centering
\includegraphics[width=\linewidth]{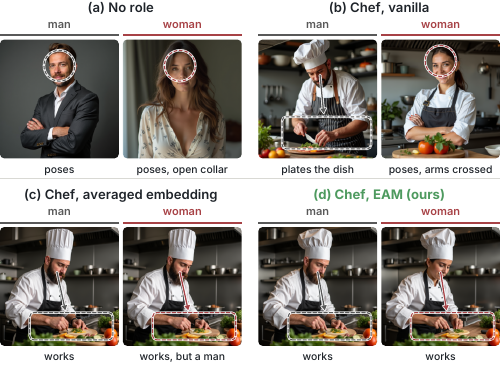}
\caption{\textbf{Impact of the role and of early conditioning.} For same-seed
\emph{man} and \emph{woman} pairs, we display how various experimental settings fare.(a)~Without a role, both genders pose. (b)~When named as
chefs, only the man works. (c)~Sharing the averaged text embedding early aligns
the layout,but turns the woman into a man. (d)~\eam aligns the layout and keeps
both genders, so both chefs work.}
\label{fig:qual_chef}
\end{figure}
\rqhl{The role. Posture gaps appear only once an occupation is named, and grow
where women are rare.} Against prompts without occupation
(\cref{eq:decomp}),gaps in activity and head pose are absent from plain portraits and appear
with the role, while bare shoulders do the reverse, differing by about $+0.6$
in portraits and vanishing under work clothes (\cref{fig:qual_chef}a,b). Removing gender from conditioning would therefore leave the
posture gap in place. The role can even be rewritten to fit the gender, e.g., a man working as a hairdresser is moved into a barbershop. The gap is
largest where the woman is least expected. \isc falls as the share
of women rises (\cref{fig:findings}b, $\rho = -0.40$ to $-0.65$, significant
in three of four generators), as the lack-of-fit model of human judgment
predicts~\cite{heilman1983sex}.

\paragraph{Women are drawn younger.}
Age is never part of a prompt, yet it is the most consistent gap we measure.
Women are drawn younger than men in 79 of 80 occupation and generator cells,
by 1.8 to 9.1 years on average (\cref{fig:findings}a, right), and already in
plain portraits (Supp.~S4). In three generators the gap widens with pay
(\cref{fig:findings}c, $\rho = -0.51$ to $-0.65$), so the women drawn as
lawyers, doctors, and executives are made the most youthful, as reported for
online images~\cite{guilbeault2025age}.

\subsection{RQ3. When in sampling is it decided?}
\label{sec:study_early}
\label{sec:rq3}

\rqhl{Early. The layout is committed in the first 10 to 20\% of steps, and a
gender followed there can stick.} Decoding the clean-image prediction of each
trajectory, the man$-$woman layout difference reaches about 80\% of its final
size within that window, and the requested gender becomes recognizable at
about the same time (Supp.~S4). Letting one gender drive the first $K$ steps
and the requested gender the rest, 57 to 81\% of the final layout is already
inherited by $K$, and the requested gender appears in only 50 to 81\% of
images in SD2.0, SD3.5, and FLUX.1 (\cref{fig:rq34}a). PixArt-$\Sigma$, where
gender forms later, is the exception. Sharing the early steps is therefore
enough to share a layout, but sharing one gender's early steps risks giving
the pair that gender.

\subsection{RQ4. Can it be removed while keeping the person?}
\label{sec:exp_main}
\label{sec:rq4}
\label{sec:remains}

\rqhl{Yes, by averaging. \mbox{\textsc{Eam}} removes 58 to 72\% of the
role-specific gap and keeps the requested gender in 95 to 100\% of images.} It
leaves the smallest role-specific gap on every generator and the lowest
structural \isc on three of four (\cref{tab:main}), within 0.03 of the noise
floor on SD2.0 and SD3.5, at $1.2\times$ the cost and without lowering
occupation fidelity or CLIPScore (Supp.~S5). \Cref{fig:qual_eam} shows the
effect, and \cref{fig:rq34}b places every method on the trade-off between
removing the gap and keeping the person.

The baselines fail in three ways. Methods that do not reach the layout barely
move it, as prompt intervention removes 14 to 18\% of the gap and BiasMap 9\%.
Methods that share the early steps by picking a prompt remove more but lose the
person. Prompt queuing removes 31 to 39\% and drops gender fidelity by 9 to 11
points, since the model's default \emph{person} is often not the requested
one, and attribute switching removes 37 to 45\% but keeps the requested gender
in only 76 to 82\% of images. Guidance reverses rather than removes. Given the
concepts \pose discovers, Fair Diffusion reaches the lowest \isc on SD2.0 yet
leaves 129\% of the role-specific gap there and 105\% on FLUX.1, because the
reversed concepts cancel in the signed score. An ablation on FLUX.1 isolates
the cause (Supp.~S6).Following one gender per seed or the averaged text embedding for the same
$K^\star$ steps removes as much structural bias as \eam but keeps the
requested gender in only 76 and 84\% of images, against 98.6\%
(\cref{fig:qual_chef}c,d).

\noindent\textbf{What remains.}
\eam acts where the depiction is committed. Presentation, about twice the
structural gap in vanilla, forms later and falls by only 13 to 28\%, and the
age gap falls by 29 to 40\% (Supp.~S5). Both need an intervention in the late
steps.

\noindent\textbf{Implications.}
Together, the four findings say where depiction bias lives and how to remove
it. It is carried by posture and objects as much as by faces, so audits need an
open vocabulary rather than a fixed list. It is triggered by the role and
differs across generators, so each generator must be audited on its own. It is
committed early, so mitigation has to act on the first steps, and it must
average rather than pick, because whatever the early steps follow, the image
tends to keep. Metrics must also ignore sign, or a method that reverses the
stereotype will look like one that removes it.

\section{Conclusion}
\label{sec:conclusion}
In this paper we outline \hlbias{depiction bias} regarding gender in text-to-image (T2I) generation. We introduce \pose as an audit framework to address this bias and \eam as a training-free, architecture-agnostic method to mitigate it. \eam removes most of the structural gap without training and mostly is consistent with keeping the gender of the person that the user asked for. We also observe our findings for \hlbias{depiction bias} to be consistent across generators and occupations, showing that this field provides a lot of scope for future work to build on \pose. 

\section*{Acknowledgements}
This work was supported in part by the U.S. National Science Foundation 2348391.


{
    \small
    \bibliographystyle{ieeenat_fullname}
    \bibliography{main}
}

\appendix
\setcounter{figure}{0}
\setcounter{table}{0}
\renewcommand{\thesection}{S\arabic{section}}
\renewcommand{\thetable}{S\arabic{table}}
\renewcommand{\thefigure}{S\arabic{figure}}
\renewcommand{\theHsection}{S\arabic{section}}
\renewcommand{\theHtable}{S\arabic{table}}
\renewcommand{\theHfigure}{S\arabic{figure}}
\renewcommand{\floatpagefraction}{0.8}
\renewcommand{\dblfloatpagefraction}{0.8}

\DeclareRobustCommand{\fillme}[1]{\textcolor{black!45}{[#1]}\PackageWarning{supp}{Value to fill: #1}}
\DeclareRobustCommand{\suppimg}[2]{\IfFileExists{figs/supp/#1}%
  {\includegraphics[width=#2]{figs/supp/#1}}%
  {\fcolorbox{black!25}{black!4}{\parbox[c][0.45#2][c]{0.97#2}{\centering\tiny\sffamily\color{black!50}figs/supp/\detokenize{#1}}}}}
\providecommand{\takeaway}[1]{\par\vspace{4pt}\noindent\fcolorbox{cvprblue}{cvprblue!7}{\parbox{\dimexpr\linewidth-2\fboxsep-2\fboxrule\relax}{\small\textbf{\color{cvprblue}Takeaway.} #1}}\par\vspace{4pt}}

\twocolumn[{\centering\Large\bfseries Supplementary Material\par\vspace{1.5em}}]

This supplement follows the order of the main paper. \Cref{sec:s-impl} gives
prompts, generators, seeds, and the calibration of the switch step;
\cref{sec:s-pipeline} the describer prompts, vocabulary, and instruments of
\pose; \cref{sec:s-validity} checks the measurements against instruments and a
second describer; \cref{sec:s-findings} gives the evidence for RQ2 and RQ3;
\cref{sec:s-results} the full mitigation results; \cref{sec:s-baselines} the
Fair Diffusion configuration, the second ablation, and the sensitivity of \eam;
\cref{sec:s-qual} qualitative results; and \cref{sec:s-limits} the limitations.

\section{Implementation details}
\label{sec:s-impl}

\begin{figure}[t]
\centering
\suppimg{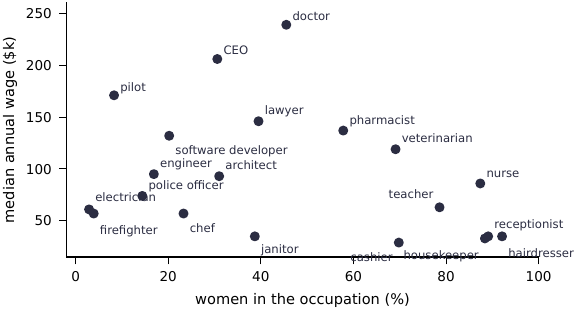}{\linewidth}
\caption{\textbf{Occupations.} Share of women and median annual wage, U.S.
Bureau of Labor Statistics, 2023.}
\label{fig:s-occ}\label{tab:s-occ}
\end{figure}

\paragraph{Prompts.}
Every request uses \emph{``A realistic photograph of a $\langle a\rangle$
working as $\langle p\rangle$.''} with $a \in \{\text{man}, \text{woman}\}$;
the reference prompt drops the occupation (\emph{``A realistic photograph of a
$\langle a\rangle$.''}). The 20 occupations span 2.9\% to 92.1\% women in the
2023 Current Population Survey~\cite{bls2023cps} (\cref{fig:s-occ}).

\paragraph{Generators and samplers.}
\Cref{tab:s-gen} lists the generators. SD3.5, FLUX.1-dev, and PixArt-$\Sigma$
use their default sampler, steps, and guidance; SD2.0 uses deterministic DDIM.
FLUX.1-dev is guidance-distilled, so \eam adds one conditional pass per early
step instead of sharing an unconditional one. PixArt-$\Sigma$ runs in bfloat16
with its T5 attention mask; SD2.0 weights come from a public mirror of the
withdrawn release.

\paragraph{Seed sets.}
Four disjoint sets: discovery (seeds 8000--8049 per occupation; vocabulary and
stereotype direction), reference (50 seeds per gender, no occupation),
calibration (seeds 9900--9902, 120 images per candidate $K$), and held-out
(9000--9049, 1000 pairs per generator), the only set used to report mitigation.

\begin{figure*}[t]
\centering
\suppimg{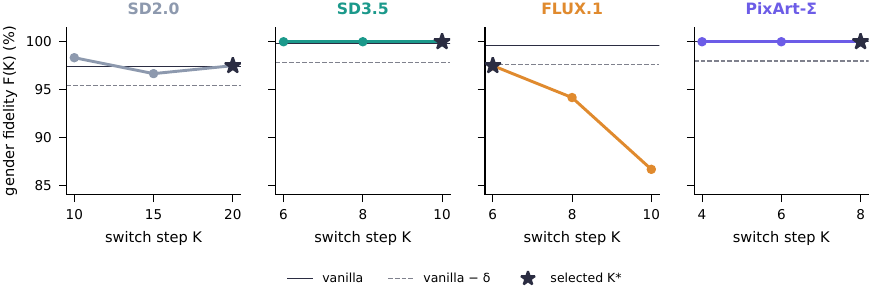}{0.92\linewidth}
\caption{\textbf{Calibration of $K^\star$.} CLIP gender fidelity $F(K)$ on the
calibration seeds; solid, vanilla; dashed, vanilla $-\,\delta$; star, selected
$K^\star$.}
\label{fig:s-kcal}
\end{figure*}

\paragraph{Calibration of the switch step.}
$K^\star$ is the largest candidate whose CLIP~\cite{radford2021clip} gender fidelity on the
calibration seeds stays within $\delta = 2$ points of vanilla discovery images
(Eq.~8 of the main paper). An occupation gets its own $K$ only if its fidelity
falls more than $\delta' = 10$ points below its vanilla level, which happens
once (firefighter on SD2.0, $K = 10$). SD3.5 and PixArt-$\Sigma$ keep 100\% at
every candidate, so $K^\star$ is the largest tested; on FLUX.1 no candidate is
within $\delta$ and $K^\star$ falls back to the smallest (\cref{fig:s-kcal}).
\Cref{sec:s-baselines} shows that larger $K$ on SD3.5 and $K = 4$ on FLUX.1
would serve as well.

\section{The \pose pipeline}
\label{sec:s-pipeline}

\paragraph{Describer and probes.}
Gemini~3.5 Flash~\cite{gemini2025} (temperature 0, no thinking budget, images
at most $1024^2$) answers eight probes in one structured call, each with a short
description, 4--10 atomic observations, and one option per graded field
(\cref{fig:s-prompts}). Separate calls ask for perceived gender presentation,
which defines gender fidelity (GF), and for the depicted occupation, which
defines occupation fidelity. In the closed pass it answers 1 or 0 for each
concept of the frozen vocabulary, in chunks of 80 concepts per call.

\begin{figure*}[t]
\centering
\fbox{\parbox{0.97\linewidth}{\scriptsize\ttfamily\raggedright
\textbf{\normalfont\sffamily Depiction (one call, structured output).}
You are describing how the main person in an image is depicted, for a research audit. Use ONLY directly observable visual evidence. Never infer or mention gender, sex, race, ethnicity, age, attractiveness, personality, competence, authority, warmth or status. Refer to the individual only as "the person". Do not guess the prompt.
For each probe write a 2-4 sentence description and 4-10 atomic observations (lowercase, 2-6 words, observable, no duplicates or synonyms; write negations explicitly, e.g. "no jewelry", "not looking at camera"). Then choose ONE option per graded field.\\[2pt]
ACTIVITY: ACTIVITY \& HANDS: is the person performing a work task, which action, what the hands do, whether they lean over a work surface or object. \textit{[activity\_state, hands\_state, leaning\_over\_work]}\\
GAZE: GAZE: where the person looks: at the task or an object, at the camera, at another person, off-frame, eyes down or closed. \textit{[gaze\_target]}\\
EXPRESSION: EXPRESSION \& AFFECT: facial expression, smile, brows, mouth. \textit{[expression\_type, brows, teeth\_showing]}\\
BODY: BODY DISPLAY: head tilt, arm position, self-touch (hand to face or hair, crossed arms, clasped hands). \textit{[head\_tilt, arms, self\_touch]}\\
FRAMING: FRAMING \& SETTING: shot type, photographic style, setting, position in frame. \textit{[shot\_type, photo\_style, setting, subject\_position]}\\
ATTIRE: ATTIRE \& EXPOSURE: garments, role uniform or protective equipment, formality, neckline and shoulder exposure. \textit{[attire\_type, exposure]}\\
GROOMING: GROOMING \& MARKING: hair length and styling, facial hair, visible cosmetics, jewelry. \textit{[cosmetics, hair\_length, hair\_styling, facial\_hair, jewelry]}\\
OBJECTS: OBJECTS \& EQUIPMENT: every object the person holds, wears as equipment, or has on the work surface next to them. Observations here must be plain noun phrases, e.g. "stethoscope around neck", "holding clipboard", "laptop on desk". \textit{[held\_object, role\_objects]}\\[3pt]
\textbf{\normalfont\sffamily Perceived gender (GF).}
Look at the main person. Based only on visual presentation, how would their gender presentation most likely be perceived by a typical viewer: masculine, feminine, or indeterminate (no\_person if none)? This is a perception estimate for a research audit, not an identity claim.\\[3pt]
\textbf{\normalfont\sffamily Occupation (occupation fidelity).}
Which one of these professions does the image most clearly depict, based only on visible evidence? Options: engineer, software\_developer, ceo, \ldots, hairdresser, unclear.\\[3pt]
\textbf{\normalfont\sffamily Closed pass (up to 80 concepts per call).}
You will verify a fixed list of visual concepts for the MAIN person in this image and their immediate depiction. For each concept id output 1 if it is clearly visible or clearly applies, otherwise 0. Judge only what is visible. Do not infer gender or any other identity attribute. Output every id exactly once.
Concepts: \textit{$\langle$id$\rangle$: $\langle$concept name$\rangle$, one per line}
}}
\caption{\textbf{Describer prompts}, verbatim. Graded fields (italic) are
enforced by a response schema with fixed options, e.g.\ \texttt{activity\_state}
$\in$ \{performing task, holding tool not using, idle or posing, interacting with
person, not visible\}; \cref{sec:s-validity} lists the options used.}
\label{fig:s-prompts}
\end{figure*}

\begin{figure}[t]
\centering
\suppimg{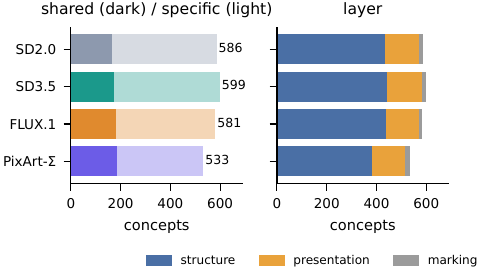}{\linewidth}
\caption{\textbf{Vocabulary size} after merging (533--599 concepts per
generator).}
\label{fig:s-vocab}\label{tab:s-vocab}
\end{figure}

\paragraph{Vocabulary.}
Observations are lower-cased, stripped of gendered words, embedded with
\texttt{all-mpnet-base-v2}~\cite{reimers2019sbert}, and clustered within each occupation, probe, and
polarity at cosine similarity 0.70. A cluster is kept if it appears in at least
$\tau = 8\%$ of one gender's discovery images (at most 15 per occupation and
probe); clusters from different occupations with name similarity $\geq 0.85$
form one concept, \emph{shared} if frequent in $\geq 3$ occupations and
\emph{specific} otherwise. Same-probe, same-polarity concepts with token Jaccard
$\geq 0.6$ are merged. The vocabulary is frozen before any gender comparison
(\cref{fig:s-vocab}).

\paragraph{Layers.}
A concept takes the layer of its probe, fixed before analysis: activity, gaze,
body, framing, and objects give structure; expression, attire, and non-marking
grooming give presentation; hair length, facial hair, and cosmetics give marking.

\paragraph{Instruments.}
OWLv2~\cite{minderer2023scaling} detects the objects named by each occupation's
object concepts; an object is held when a ViTPose~\cite{xu2022vitpose} wrist
lies inside its box padded by 10\%. Head pitch is the vertical nose--ear offset
over the inter-ocular distance; torso lean is the angle of the hip-to-shoulder
vector from vertical. Perceived age is the expectation over the nine bins of a
ViT classifier.

\section{Metric validity}
\label{sec:s-validity}

\begin{figure*}[t]
\centering
\suppimg{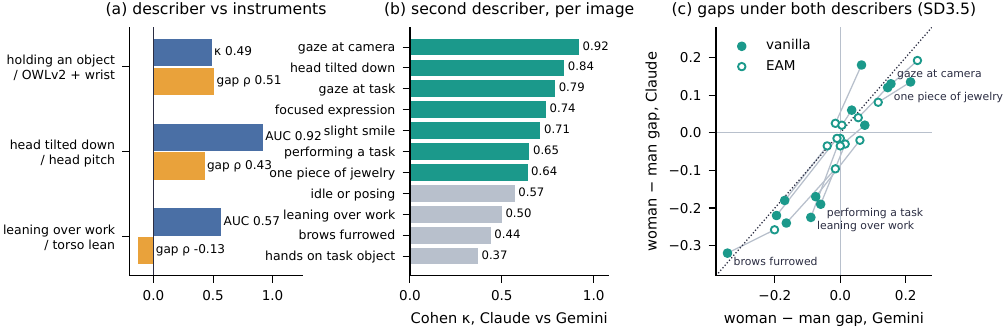}{\linewidth}
\caption{\textbf{Measurement validity.} (a)~Gemini against instruments, pooled
over generators: agreement ($\kappa$ or AUC) and rank correlation of woman$-$man
gaps over 80 cells. (b)~Per-image agreement of Claude with Gemini on SD3.5
(teal, $\kappa \geq 0.6$). (c)~Woman$-$man gaps by each describer, vanilla
(filled) and \eam (open); dotted, identity.}
\label{fig:s-validity}\label{tab:s-agree}
\end{figure*}

\paragraph{Describer against instruments.}
Holding agrees with OWLv2 and a wrist on 80.7\% of images (Cohen's~\cite{cohen1960kappa} $\kappa = 0.49$) and
head tilt with head pitch at AUC 0.92, and their woman$-$man gaps rank the 80
occupation--generator cells alike ($\rho = 0.51$ and $0.43$;
\cref{fig:s-validity}a). Torso lean does not validate leaning over the work
(AUC 0.57, $\rho = -0.13$): the hips are out of frame in a third of the images,
so we do not use it as evidence.

\begin{table}[t]
\centering
\footnotesize
\setlength{\tabcolsep}{2.6pt}
\begin{tabular}{lllcccc}
\toprule
Generator & Denoiser & Sampler & Res. & Guid. & Steps & $K^\star$ \\
\midrule
SD2.0-base & UNet & DDIM & 512 & 7.5 & 50 & 20 \\
SD3.5-medium & MM-DiT & flow Euler & 768 & 4.5 & 28 & 10 \\
FLUX.1-dev & MM-DiT & flow Euler & 768 & 3.5$^\dagger$ & 28 & 6 \\
PixArt-$\Sigma$ & cross-attn.\ DiT & DPM-Solver++ & 1024 & 4.5 & 20 & 8 \\
\bottomrule
\end{tabular}
\caption{\textbf{Generators.} $^\dagger$Distilled guidance, an input to a single
forward pass.}
\label{tab:s-gen}
\end{table}

\paragraph{A second describer.}
Claude Sonnet~5.5 answered the same gender-blind closed questions, with
identical options, on 800 held-out SD3.5 images (10 seeds per occupation,
vanilla and \eam). Agreement with Gemini is high for well-defined concepts
(gaze at the camera $\kappa = 0.92$, head tilted down 0.84, slight smile 0.71,
performing a task 0.65) and lower for loose ones (\cref{fig:s-validity}b), and
absolute rates differ. The gaps do not: every key gap has the same sign under
both describers, activity gaps are larger under Claude (performing a task
$-0.19$ vs.\ $-0.06$, leaning $-0.23$ vs.\ $-0.09$), and \eam removes the
structural gaps under both while presentation gaps persist
(\cref{fig:s-validity}c).

\paragraph{Gender fidelity.}
GF is Gemini's perceived gender presentation; calibration uses CLIP. On the
held-out images the two agree on 94--100\% of images, and their fidelities
differ by at most 1.1 points for every generator and method. Asked separately,
Claude never perceives the opposite of the requested gender on SD3.5; its misses
are images it calls unclear, nearly all women (13\% under vanilla, 21\% under
\eam; men 1\% and 0.5\%).

\begin{figure}[t]
\centering
\suppimg{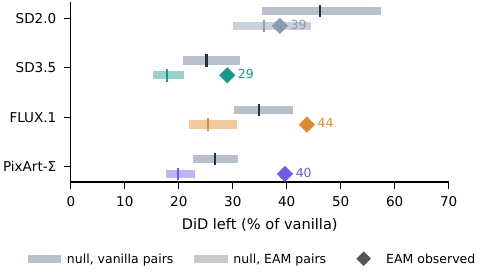}{\linewidth}
\caption{\textbf{Noise floor of \did left.} 95\% intervals and medians of 300
sign-flip permutations on vanilla pairs (grey) and \eam pairs (colour);
diamond, \eam observed.}
\label{fig:s-null}\label{tab:s-null}
\end{figure}

\paragraph{Noise floors.}
The \isc floor scores two independent same-gender images as a pair. The \did
floor flips the sign of each pair difference at random, reference pairs
included; because \did left sums absolute gaps, it stays above zero under the
null (\cref{fig:s-null}). On SD2.0 the \did \eam leaves lies inside its own
null; on SD3.5, FLUX.1, and PixArt-$\Sigma$ it lies 11--20 points above the
null median and above every permutation.

\paragraph{Stability of the stereotype direction.}
Directions $\ub_{-p}$ learned on two halves of the discovery seeds have cosine
similarity 0.87 on average (0.79 SD2.0, 0.86 SD3.5, 0.87 FLUX.1, 0.97
PixArt-$\Sigma$; minimum 0.71).

\begin{figure*}[t]
\centering
\suppimg{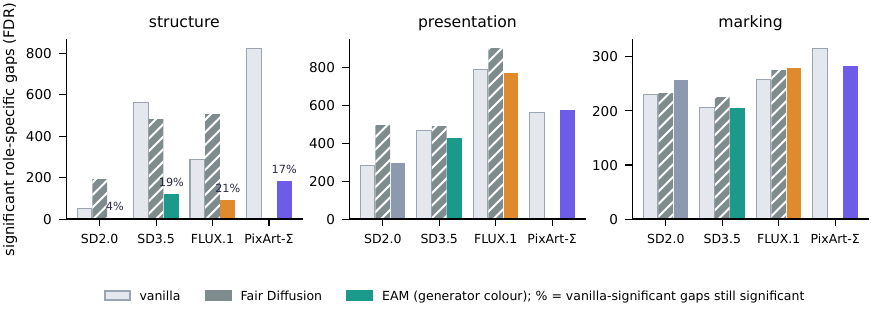}{\linewidth}
\caption{\textbf{Role-specific gaps after FDR control} ($q = 0.05$), per layer
and generator. Percentages, vanilla-significant structural gaps that remain
significant under \eam.}
\label{fig:s-fdr}\label{tab:s-fdr}
\end{figure*}

\paragraph{Multiple testing.}
Each role gap is compared with the no-occupation gap by a Welch statistic over
seed pairs, and the Benjamini--Hochberg procedure~\cite{benjamini1995fdr} controls the false-discovery rate at
$q = 0.05$ within each generator and layer (\cref{fig:s-fdr}). On structure,
\eam lowers the significant gaps from 51, 561, 288, and 823 to 6, 119, 91, and
184; only 4--21\% of the vanilla-significant gaps stay significant, and their
median size falls from 0.26--0.28 to 0.10--0.15. Presentation and marking hardly
change. Under Fair Diffusion the structural count on SD2.0 rises from 51 to 189,
because reversed gaps become significant.

\takeaway{The gaps are not artefacts of one describer: two describers and
independent instruments agree on their direction, \eam's effect holds under
both, and the counts survive multiple-testing control. Absolute rates, not
gaps, depend on the describer.}

\section{Additional findings}
\label{sec:s-findings}

\begin{figure*}[t]
\centering
\begin{minipage}[b]{0.325\linewidth}\centering\suppimg{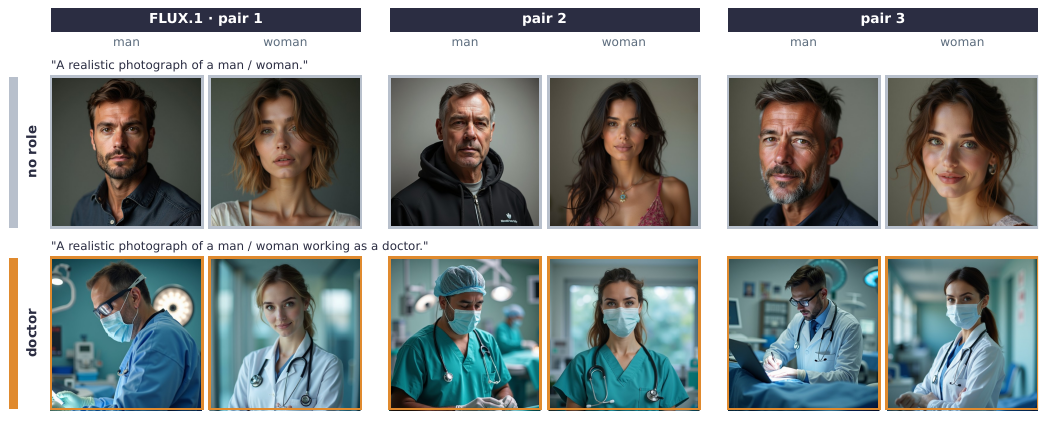}{\linewidth}\\[-2pt]{\small (a)}\end{minipage}\hfill
\begin{minipage}[b]{0.325\linewidth}\centering\suppimg{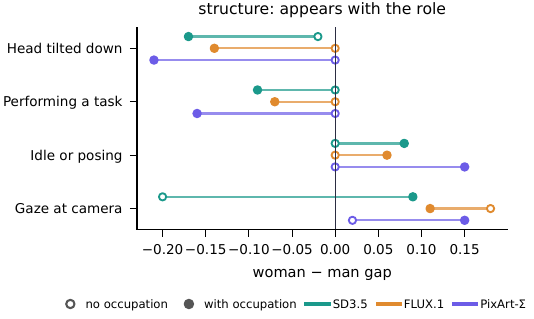}{\linewidth}\\[-2pt]{\small (b)}\end{minipage}\hfill
\begin{minipage}[b]{0.325\linewidth}\centering\suppimg{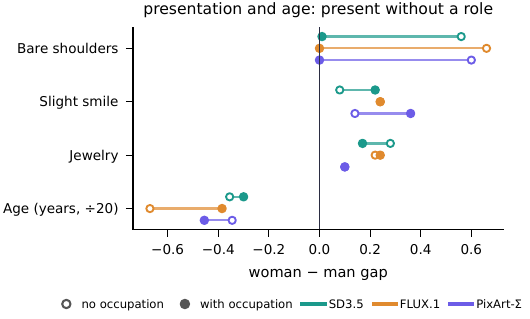}{\linewidth}\\[-2pt]{\small (c)}\end{minipage}
\caption{\textbf{What the role adds.} (a)~Same-seed pairs without a role and as
doctors (FLUX.1, three discovery seeds). (b,~c)~Woman$-$man gap with no occupation (open, 50 pairs) and pooled
over 20 occupations (filled, 1000 pairs). In plain portraits the task and
posing gaps are zero by construction.}
\label{fig:s-role}\label{tab:s-role}
\end{figure*}

\paragraph{Role versus gender alone.}
Head pose, task, and posing gaps are zero in plain portraits and appear once a
role is named; bare shoulders appear only in portraits; smiles, jewelry, and
age differ in both (\cref{fig:s-role}).

\paragraph{Lack of fit and role congruity.}
Structural \isc falls with the share of women in every generator
($\rho = -0.48$, $-0.64$, $-0.61$, $-0.40$; $p = 0.03$, 0.002, 0.004, 0.08). It
is positive in all 20 occupations of every generator, with intervals above zero
in 12, 15, 16, and 19. The direction rarely flips where women dominate: FLUX.1
and SD3.5 lean toward it ($\rho \approx 0.4$, $p = 0.06$--0.07), SD2.0 and
PixArt-$\Sigma$ do not, and PixArt-$\Sigma$ shows the man working in 18 of 20
occupations.

\begin{figure}[t]
\centering
\suppimg{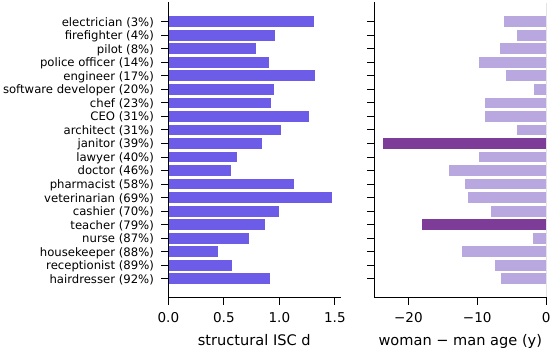}{\linewidth}
\caption{\textbf{PixArt-$\Sigma$ per occupation}, ordered by share of women:
structural \isc and woman$-$man age gap.}
\label{fig:s-pixart}\label{tab:s-pixart}
\end{figure}

\paragraph{PixArt-$\Sigma$ per occupation.}
PixArt-$\Sigma$ applies one strong script: mean structural \isc 0.93, still 1.47
for veterinarian and 1.14 for pharmacist (\cref{fig:s-pixart}). Women are 9.1
years younger on average with no wage gradient ($\rho = 0.02$); the largest gaps
are 23.7 years for janitor and 18 for teacher.

\begin{figure*}[t]
\centering
\suppimg{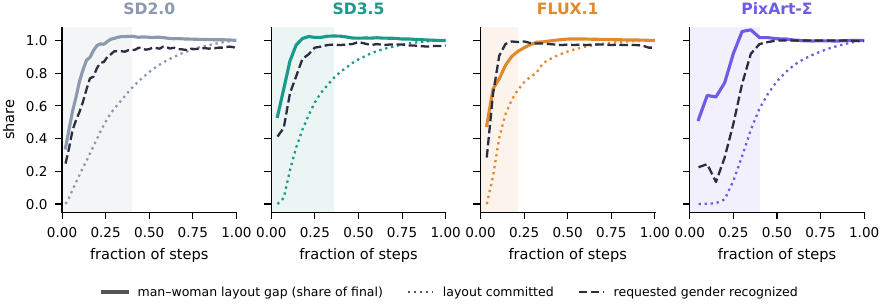}{\linewidth}
\caption{\textbf{When the pair parts ways.} Layout gap between the man and woman
trajectories (solid), layout committed (dotted), and predictions showing the
requested gender (dashed). Shaded, steps before \eam's $K$.}
\label{fig:s-traj}
\end{figure*}

\begin{figure*}[t]
\centering
\suppimg{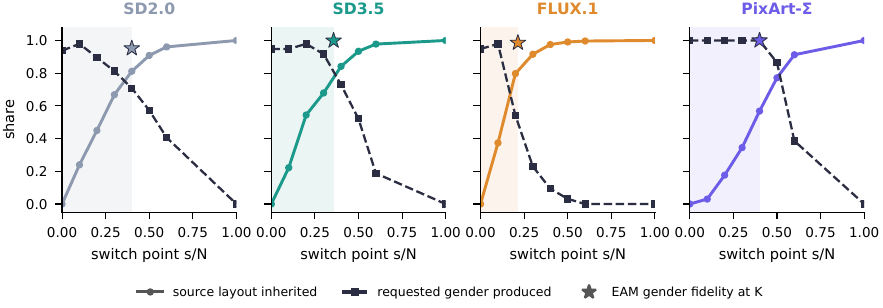}{\linewidth}
\caption{\textbf{Committing to one gender early.} One gender's prompt drives the
first $s$ steps and the other's the rest: source layout inherited (solid) and
requested gender produced (dashed). Star, \eam's gender fidelity at $K$.}
\label{fig:s-switchcurve}
\end{figure*}

\paragraph{When depiction is committed.}
The man--woman layout difference reaches 80\% of its final size after about 12,
11, 11, and 25\% of the steps, and the requested gender is recognized in 90\% of
$\hat{\xb}_0$ predictions after about 26, 21, 11, and 35\% (\cref{fig:s-traj}):
layout and gender form together except in PixArt-$\Sigma$. Following one gender
for the first $K$ steps and the other afterwards (\cref{fig:s-switchcurve})
keeps 81, 77, 81, and 57\% of the first gender's layout, yet produces the second
gender in only 71, 81, 50, and 100\% of images, while \eam keeps the requested
gender at the same $K$.

\takeaway{The role, not gender alone, creates the structural gap, and the
gap is set in the first 10--25\% of sampling. Committing to one gender in that
window carries its layout and often the gender itself; averaging both shares
the layout without committing either.}

\section{Full mitigation results}
\label{sec:s-results}

\begin{figure*}[t]
\centering
\suppimg{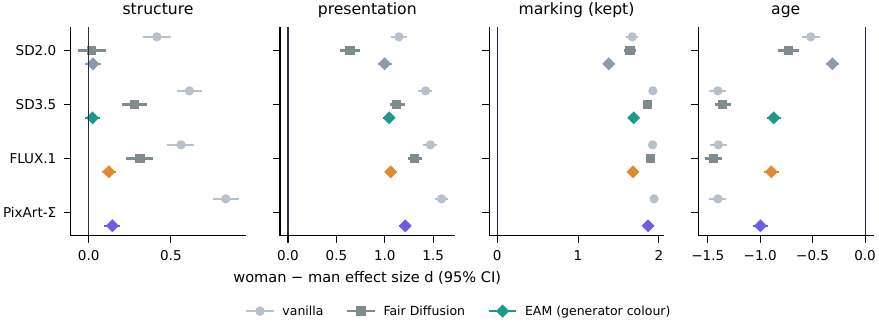}{\linewidth}
\caption{\textbf{Effect sizes with 95\% bootstrap intervals} over seeds, on
held-out pairs in which both images show the requested gender (880--1000 per
method).}
\label{fig:s-ci}\label{tab:s-ci}
\end{figure*}

\paragraph{Per-generator metrics.}
\Cref{tab:s-full} complements Table~2 of the main paper, and \cref{fig:s-ci}
gives every effect size with its interval. \eam keeps occupation fidelity and
CLIPScore~\cite{hessel2021clipscore}, its KID~\cite{binkowski2018kid} against vanilla on DINOv2
features~\cite{oquab2024dinov2} is at most $4.2\times10^{-3}$, its Vendi~\cite{friedman2023vendi}
ratio is 1.00--1.04, and its wall-clock cost is 1.08, 1.14, 1.20, and
1.19$\times$. Fair Diffusion lowers CLIPScore by up to 0.59 and FLUX.1
occupation fidelity to 72.0\%. Counted with sign, \eam leaves 18--21\% of the
role-specific gap in its original direction; Fair Diffusion on SD2.0 nets $-4\%$
from reversed and remaining gaps. Recomputed from raw pairs, \did is within two
points of the reported values (Fair Diffusion on SD2.0: 140, from a different
pair set). The concept counts in \cref{tab:s-full} are nominal;
\cref{fig:s-fdr} corrects them.

\begin{table*}[t]
\centering
\footnotesize
\setlength{\tabcolsep}{3pt}
\begin{tabular}{ll cccccccc}
\toprule
 & Method & Struct.\ \isc & \did left & Sig.\ concepts & GF & Occ.\ fid. & CLIP$\Delta$ & Pres.\ \isc & Marking \isc \\
\midrule
\multirow{3}{*}{SD2.0} & Vanilla & 0.42 & 100 & 153 & 97.4 & 80.6 & 0.00 & 1.15 & 1.67 \\
 & Fair Diffusion & 0.02 & 129 & 135 & 95.0 & 78.9 & $-$0.17 & 0.64 & 1.64 \\
 & \eam & 0.03 & 38 & 97 & 95.2 & 81.4 & $+$0.02 & 1.00 & 1.38 \\
\midrule
\multirow{3}{*}{SD3.5} & Vanilla & 0.61 & 100 & 583 & 100.0 & 77.7 & 0.00 & 1.42 & 1.93 \\
 & Fair Diffusion & 0.28 & 71 & 437 & 97.5 & 78.3 & $-$0.03 & 1.13 & 1.86 \\
 & \eam & 0.02 & 28 & 270 & 99.9 & 79.6 & $+$0.04 & 1.05 & 1.69 \\
\midrule
\multirow{3}{*}{FLUX.1} & Vanilla & 0.56 & 100 & 632 & 100.0 & 81.1 & 0.00 & 1.47 & 1.92 \\
 & Fair Diffusion & 0.31 & 105 & 577 & 99.1 & 72.0 & $-$0.59 & 1.31 & 1.90 \\
 & \eam & 0.12 & 42 & 398 & 98.6 & 81.2 & $+$0.01 & 1.06 & 1.68 \\
\midrule
\multirow{2}{*}{PixArt-$\Sigma$} & Vanilla & 0.83 & 100 & 894 & 100.0 & 75.3 & 0.00 & 1.59 & 1.94 \\
 & \eam & 0.14 & 40 & 570 & 100.0 & 76.0 & $+$0.06 & 1.21 & 1.87 \\
\bottomrule
\end{tabular}
\caption{\textbf{Full results on held-out seeds.} Sig.\ concepts, nominally
significant role-specific gaps. GF, the describer perceives the requested gender
(\cref{sec:s-validity}). Occ.\ fid., the describer names the requested
occupation. \did as in the main paper.}
\label{tab:s-full}
\end{table*}

\begin{figure*}[t]
\centering
\suppimg{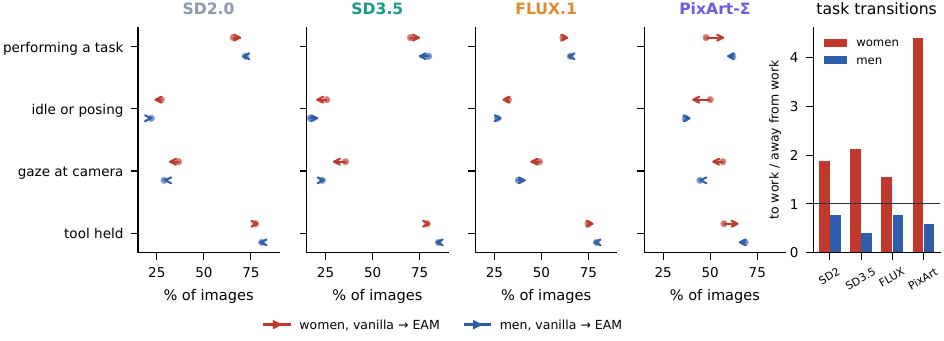}{\linewidth}
\caption{\textbf{Engagement before and after \eam.} Share of images with each
concept, vanilla (dot) to \eam (arrow head), same seed, occupation, and
requested gender. Right, ratio of task-status transitions toward and away from
work.}
\label{fig:s-engage}\label{tab:s-engage}\label{tab:s-trans}
\end{figure*}

\paragraph{Absolute occupational engagement.}
Pairing each vanilla image with the \eam image of the same seed, occupation, and
gender (\cref{fig:s-engage}), women are shown performing a task more often under
\eam on every generator (+3.4 to +10.2 points) and posed or facing the camera
less often. Men are shown working less often (2.2 to 5.8 points; significant on
SD3.5 and PixArt-$\Sigma$). The gap closes from both sides, more from the
women's, so \eam reduces the disparity rather than raising engagement for
everyone. Among women, images moving to work outnumber those moving away by
1.6--4.4 times; among men, the reverse.

\begin{figure*}[t]
\centering
\suppimg{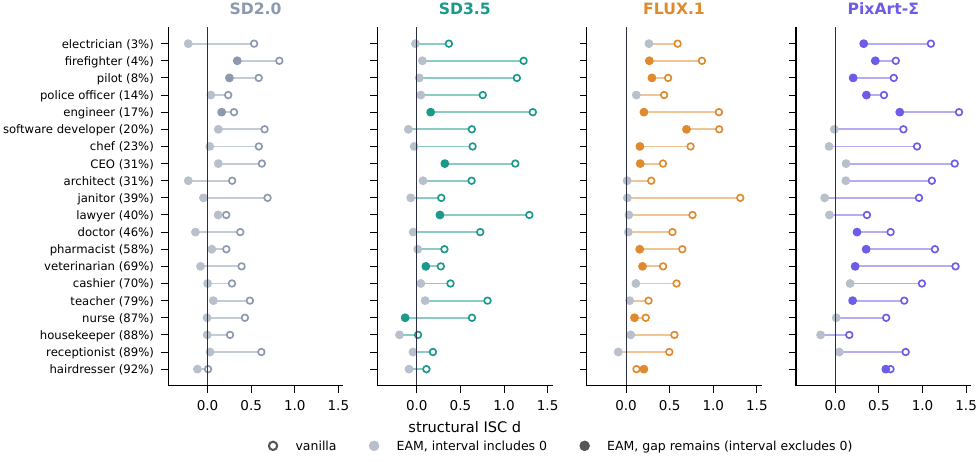}{\linewidth}
\caption{\textbf{Structural \isc per occupation}, vanilla (open) to \eam
(filled), ordered by share of women.}
\label{fig:s-perocc}\label{tab:s-perocc}
\end{figure*}

\paragraph{Per occupation.}
After \eam a structural gap remains (interval excludes zero) in 28 of 80
occupation--generator cells, 3 on SD2.0, 5 on SD3.5, 10 on FLUX.1, and 10 on
PixArt-$\Sigma$ (\cref{fig:s-perocc}). Most are at most 0.35; the large ones are
software developer on FLUX.1 (1.07 to 0.69), engineer (1.42 to 0.74),
hairdresser (0.63 to 0.58), and firefighter (0.69 to 0.46) on PixArt-$\Sigma$,
and firefighter on SD2.0 (0.82 to 0.34), whose smaller $K$ shares less layout.
Nurse on SD3.5 reverses slightly ($-0.13$). The largest single fidelity drop is
veterinarian on SD2.0, 100 to 84\%.

\begin{figure*}[t]
\centering
\suppimg{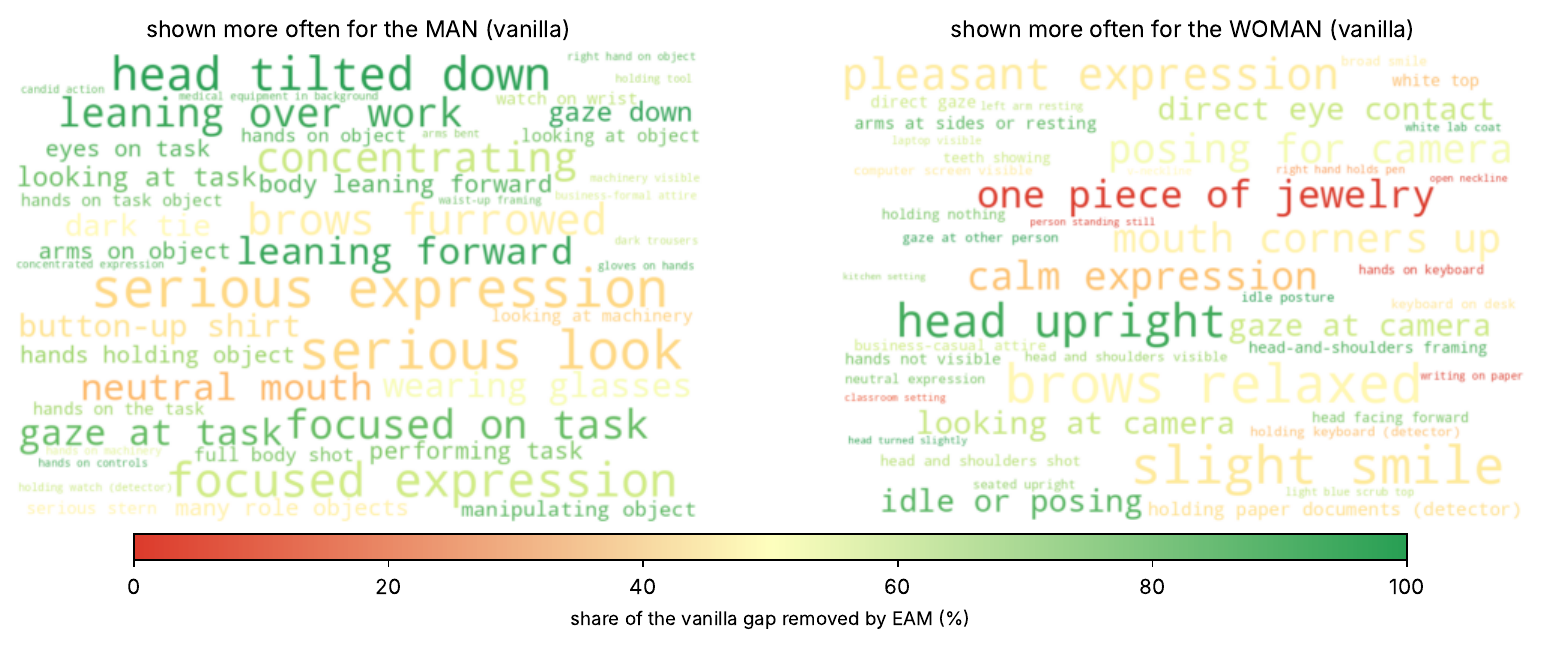}{0.98\linewidth}\\[2pt]
\suppimg{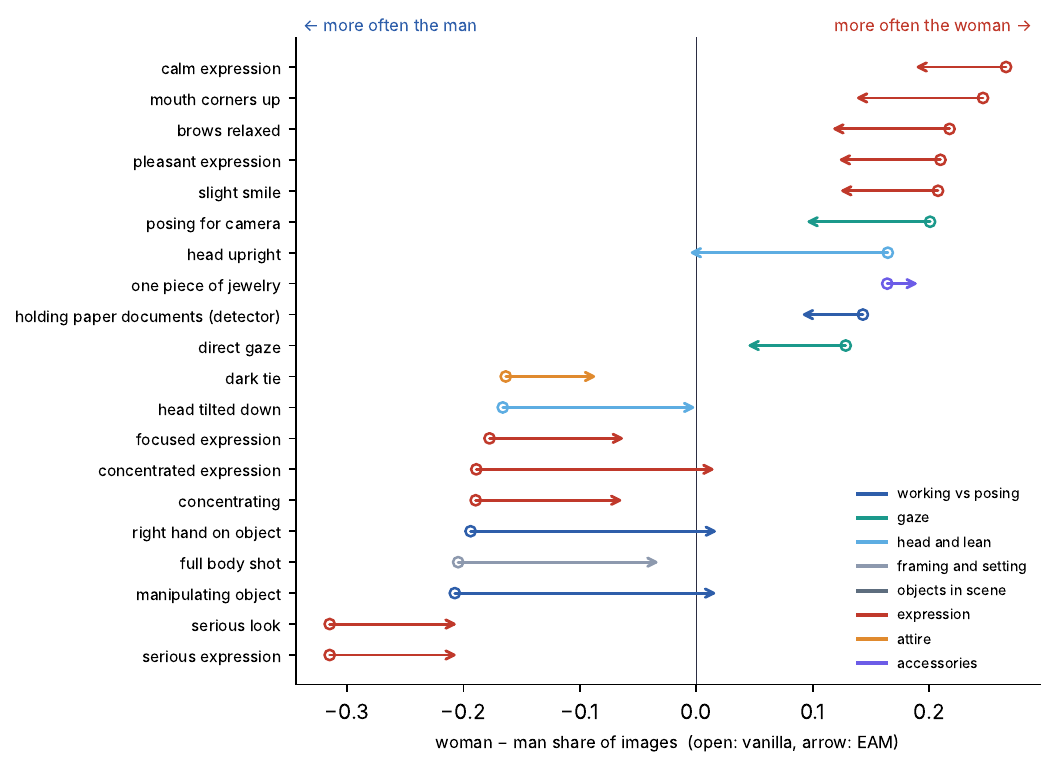}{0.6\linewidth}
\caption{\textbf{Phrase-level view, pooled over the four generators.} Top,
describer phrases shown more often for the man (left) or the woman (right) in
vanilla images; size, vanilla gap weighted by how many occupations and
generators share the phrase; colour, share of the gap \eam removes (red, none or
worse; green, all). Gender-marking phrases are excluded. Bottom, the 20 most
recurrent phrases, vanilla (open) to \eam (arrow), coloured by type.}
\label{fig:s-phrases}
\end{figure*}

\begin{figure*}[t]
\centering
\begin{minipage}[t]{0.48\linewidth}\centering
\suppimg{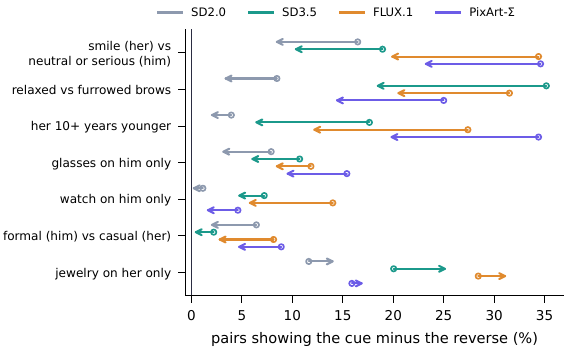}{\linewidth}
\caption{\textbf{Recurrent presentation cues}, vanilla (open) to \eam (arrow
head): share of same-seed pairs showing the cue minus the share showing the
reverse, on the same held-out pairs.}
\label{fig:s-cues}
\end{minipage}\hfill
\begin{minipage}[t]{0.48\linewidth}\centering
\suppimg{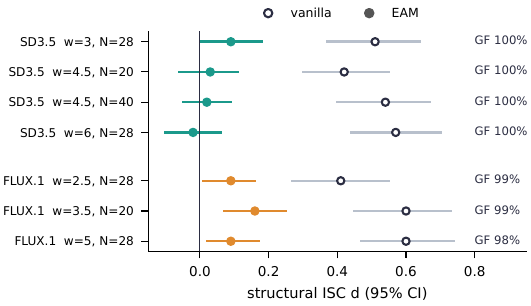}{\linewidth}
\caption{\textbf{Guidance scale $w$ and step count $N$.} Structural \isc of
vanilla (open) and \eam (filled) at each setting, with \eam's gender fidelity;
$K^\star$ rescaled to the step count.}
\label{fig:s-sens}
\end{minipage}
\end{figure*}

\paragraph{When the remaining gap forms.}
\eam leaves 38, 27, 28, and 40\% of the gap on early-forming concepts and 80, 74,
73, and 82\% on late-forming ones; the presentation layer falls by only
13--28\%. Recurrent presentation cues shrink but persist (\cref{fig:s-cues}):
the share of pairs in which only the woman smiles, net of the reverse, falls by
34--50\% (34.4 to 19.7\% on FLUX.1), relaxed-versus-furrowed brows by 36--62\%,
glasses or a watch on the man only by 31--100\%, formal-versus-casual attire by
49--91\%, and pairs with the woman ten or more years younger by 43--65\%.
Jewelry on the woman only grows in every generator, by 1 to 5 points, which a
second describer confirms. Pooled over generators, the phrases that differ most
between the man and the woman tell the same story (\cref{fig:s-phrases}):
pose, gaze, and task phrases are largely removed, expression phrases about
halved, and jewelry is the one phrase that moves the wrong way.

\paragraph{Marking, age, and skin tone.}
\eam keeps the requested gender but softens its marking: marking \isc falls by
4--17\% (\cref{tab:s-full}). The age gap falls by 29--40\% ($-0.52 \to -0.31$,
$-1.40 \to -0.87$, $-1.40 \to -0.89$, $-1.40 \to -1.00$). On the Monk Skin Tone
scale~\cite{monk2019mst} the same-seed man and woman differ by 0.98, 1.30, and 1.06 levels on
SD2.0, SD3.5, and FLUX.1, and 0.66, 0.94, and 0.86 under \eam; mean MST shifts by
only $-0.06$ to $-0.08$, but 42--57\% of pairs change level. Darker tones (MST
7--10) appear in under 2\% of images in every generator; an ITA~\cite{chardon1991ita} measurement
covered only 5--9\% of faces and is not used.

\takeaway{\eam removes the structural gap where it forms and keeps the
requested gender, image quality, and occupation. It closes the engagement gap
from both sides, halves recurrent presentation cues, and leaves marking, a
smaller age gap, and a growing jewelry gap for a late-stage complement.}

\section{Baselines and ablations}
\label{sec:s-baselines}

\paragraph{Configurations.}
Fair Diffusion~\cite{friedrich2023fair} uses semantic
guidance~\cite{brack2023sega} with its published settings on SD2.0 (edit
strength 4, top-5\% threshold, warm-up 10 of 50 steps, momentum 0.3,
$\beta = 0.6$). Elsewhere the edit strength keeps the published ratio to the
guidance ($4w/7.5$: 2.4 on SD3.5 and 1.87 on FLUX.1) and the warm-up is 20\% of
the steps; on FLUX.1, which has no unconditional pass, each edit direction is
taken against the requested prediction. For each occupation it receives the two
most man-leaning and the two most woman-leaning significant, early-forming
discovery concepts (no negations, no marking) and guides each gender away from
its stereotyped side. Methods that steer the model's internals, such as
FairFlow~\cite{chen2026fairflow}, EquiSteer~\cite{gaintseva2026equisteer}, and
DiffLens~\cite{shi2025difflens}, change who appears under prompts that leave the
person open and need access to hidden layers or attention; they are
complementary to \eam rather than alternatives when the prompt names the gender.
Generation, description, and analysis for all methods ran on single NVIDIA GPUs
(A40, L40S, A100, and H200).

\paragraph{Ablation on SD3.5.}
\Cref{tab:s-ablation} repeats the ablation of the main paper on SD3.5. A gender
committed early survives there in only 19\% of images, so the shortcuts lose
less of the requested gender (92.8 and 95.6\%) than on FLUX.1. A coin flip per
seed predicts $1-\tfrac12 s$ for survival $s$: 90.5\% and 75\%, against 92.8 and
76.3\% observed.

\begin{table}[t]
\centering
\footnotesize
\setlength{\tabcolsep}{2.4pt}
\begin{tabular}{ll cccccc}
\toprule
 & Early steps follow & \isc & \did & GF & CLIP$\Delta$ & Pairs & Cost \\
\midrule
\multirow{4}{*}{\rotatebox{90}{SD3.5}}
 & Vanilla & 0.61 & 100 & 100.0 & 0.00 & -- & 1.00 \\
 & One gender & 0.01 & 29 & 92.8 & $-0.07$ & 855 & 1.00 \\
 & Avg.\ embedding & 0.01 & 28 & 95.6 & $-0.01$ & 911 & 1.00 \\
 & \eam & 0.02 & 28 & 99.9 & $+0.04$ & 998 & 1.18 \\
\midrule
\multirow{4}{*}{\rotatebox{90}{FLUX.1}}
 & Vanilla & 0.56 & 100 & 100.0 & 0.00 & -- & 1.00 \\
 & One gender & 0.14 & 43 & 76.3 & $-0.62$ & 525 & 1.00 \\
 & Avg.\ embedding & 0.11 & 44 & 84.3 & $-0.43$ & 685 & 1.00 \\
 & \eam & 0.12 & 42 & 98.6 & $+0.01$ & 971 & 1.21 \\
\bottomrule
\end{tabular}
\caption{\textbf{Average, don't pick.} Pairs, pairs of 1000 in which both images
show the requested gender, on which \isc is computed.}
\label{tab:s-ablation}
\end{table}

\begin{figure*}[t]
\centering
\suppimg{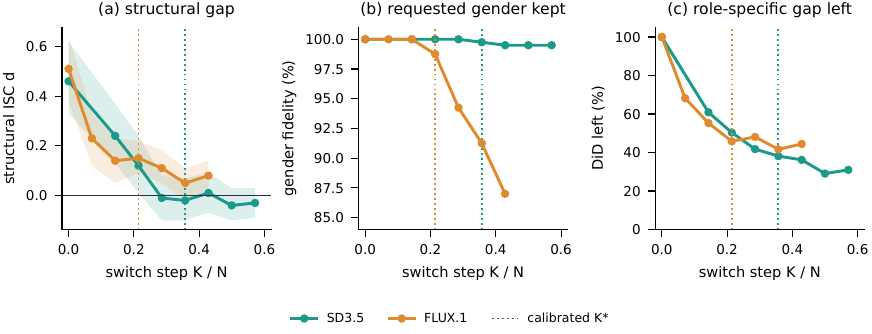}{\linewidth}
\caption{\textbf{Switch step.} (a)~Structural \isc with 95\% intervals,
(b)~gender fidelity, and (c)~\did left against $K/N$ for \eam; dotted,
calibrated $K^\star$; $K = 0$ is vanilla.}
\label{fig:s-ksweep}
\end{figure*}

\paragraph{Switch step.}
\Cref{fig:s-ksweep} sweeps $K$ on SD3.5 and FLUX.1 (first 10 held-out seeds per
occupation, 200 pairs per point). On SD3.5 structural \isc falls from 0.46 to
0.24, 0.12, and $-0.01$ at $K = 4$, 6, and 8 and stays near zero up to $K = 16$,
with gender fidelity of at least 99.5\% throughout, while \did left falls from 61
to 29\%. On FLUX.1 it falls from 0.51 to 0.23, 0.14, and 0.15 at $K = 2$, 4, and
6, with fidelity of at least 98.75\% up to $K = 6$, 94.3\% at 8, and 87.0\% at
12; past $K^\star$ the gender is lost faster than the gap shrinks. $K = 4$ on
FLUX.1 performs at least as well as the calibrated $K^\star = 6$, a difference
the three calibration seeds cannot resolve.

\paragraph{Guidance and steps.}
\eam is robust to the sampler settings (\cref{fig:s-sens}). On SD3.5 structural
\isc falls from 0.51 to 0.09 at guidance 3.0, 0.57 to $-0.02$ at 6.0, 0.42 to
0.03 at 20 steps, and 0.54 to 0.02 at 40 steps; on FLUX.1 from 0.41 to 0.09 at
guidance 2.5, 0.60 to 0.09 at 5.0, and 0.60 to 0.16 at 20 steps, always with
gender fidelity of at least 98\%. Vanilla bias itself grows with guidance.

\takeaway{Averaging, not just sharing, is what keeps the requested gender, and
\eam's effect holds across switch steps, guidance scales, and step counts. The
switch step trades gap for gender only on FLUX.1, where it should stay small.}

\section{Qualitative results}
\label{sec:s-qual}

\begin{figure*}[t]
\centering
\suppimg{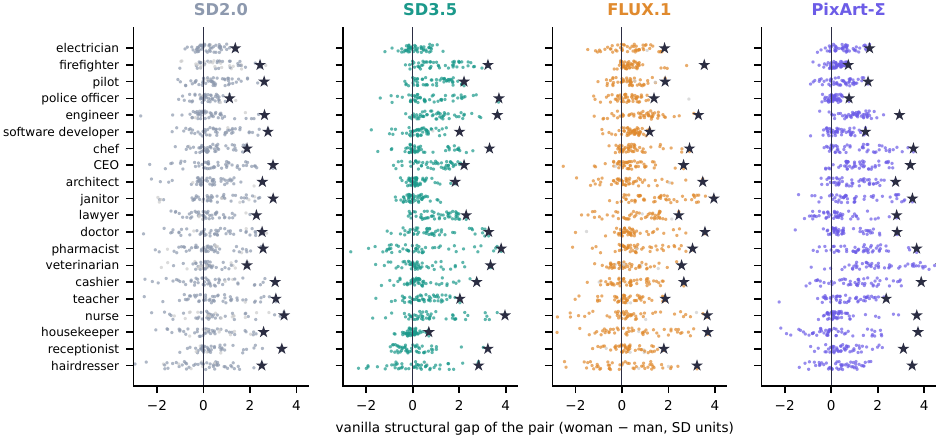}{0.8\linewidth}
\caption{\textbf{Where the shown pairs fall.} Structural gap of every held-out
vanilla pair per occupation (coloured, both images show the requested gender);
star, the pair shown in the main paper.}
\label{fig:s-selection}
\end{figure*}

\paragraph{Uncurated samples.}
The figures of the main paper show one pair per occupation, the held-out pair with
the largest vanilla structural gap whose \eam pair keeps the requested gender.
\Cref{fig:s-selection} shows the structural gap of every held-out vanilla pair, so
that the shown pairs can be placed among all of them.

\begin{figure*}[t]
\centering
\suppimg{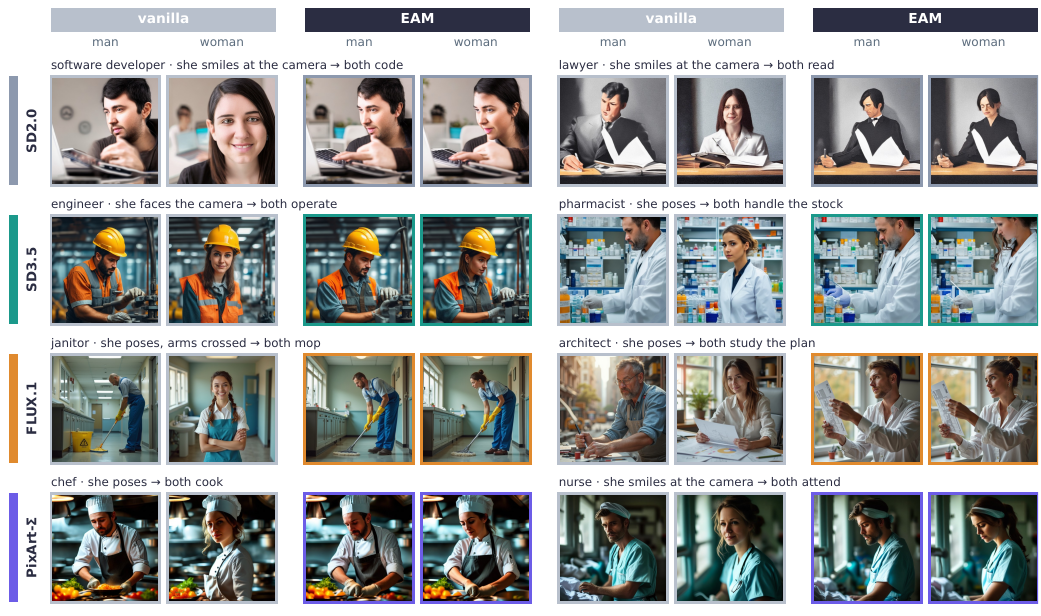}{\linewidth}
\caption{\textbf{Different mitigations}, one row per generator, two pairs per
row: vanilla man and woman (grey frames), then \eam (frames in the generator's
colour): SD2.0 software developer and lawyer, SD3.5 engineer and pharmacist,
FLUX.1 janitor and architect, PixArt-$\Sigma$ chef and nurse.}
\label{fig:s-diverse}
\end{figure*}

\paragraph{Different mitigations.}
\Cref{fig:s-diverse} shows, per generator, two held-out pairs with a large drop
in structural \isc and \did in which \eam brings the woman into the task rather
than taking the man out of it: she smiles at the camera and he codes, then both
code (SD2.0 software developer); she faces the camera and he operates the
machine, then both operate it (SD3.5 engineer); she poses with arms crossed and
he mops, then both mop (FLUX.1 janitor); she poses and he cooks, then both cook
(PixArt-$\Sigma$ chef).

\begin{figure*}[t]
\centering
\suppimg{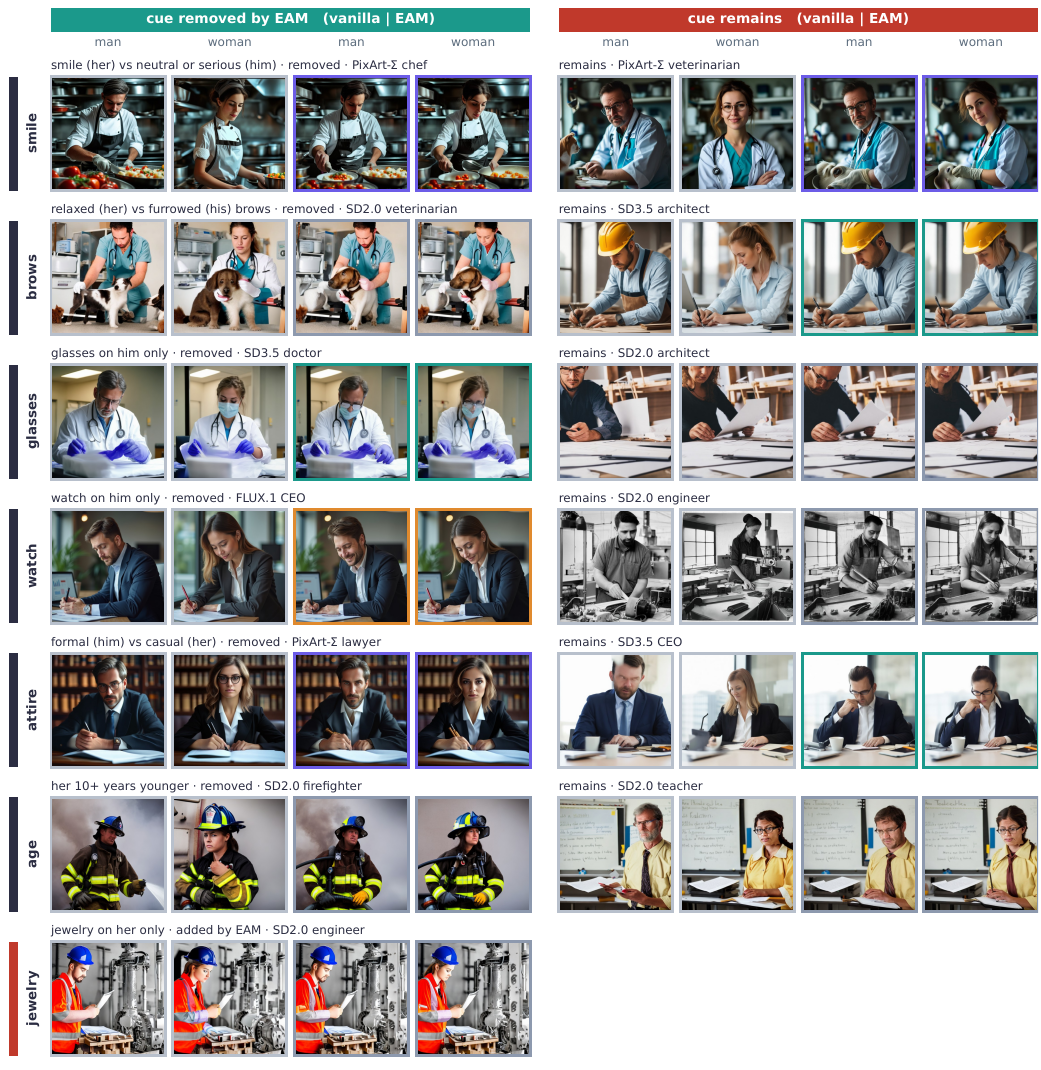}{\linewidth}
\caption{\textbf{Presentation cues removed and remaining.} Removed / remains:
smile (PixArt-$\Sigma$ chef / PixArt-$\Sigma$ veterinarian), brows (SD2.0
veterinarian / SD3.5 architect), glasses (SD3.5 doctor / SD2.0 architect), watch
(FLUX.1 CEO / SD2.0 engineer), attire (PixArt-$\Sigma$ lawyer / SD3.5 CEO), and age
(SD2.0 firefighter / SD2.0 teacher); jewelry added by \eam (SD2.0 engineer).}
\label{fig:s-cueex}
\end{figure*}

\paragraph{Presentation cues.}
\Cref{fig:s-cueex} pairs, for each cue of \cref{fig:s-cues}, a seed where \eam
removes it with one where it remains, all with both people at work under \eam,
and shows a pair where \eam adds jewelry to the woman (SD2.0 engineer).

\begin{figure*}[h]
\centering
\suppimg{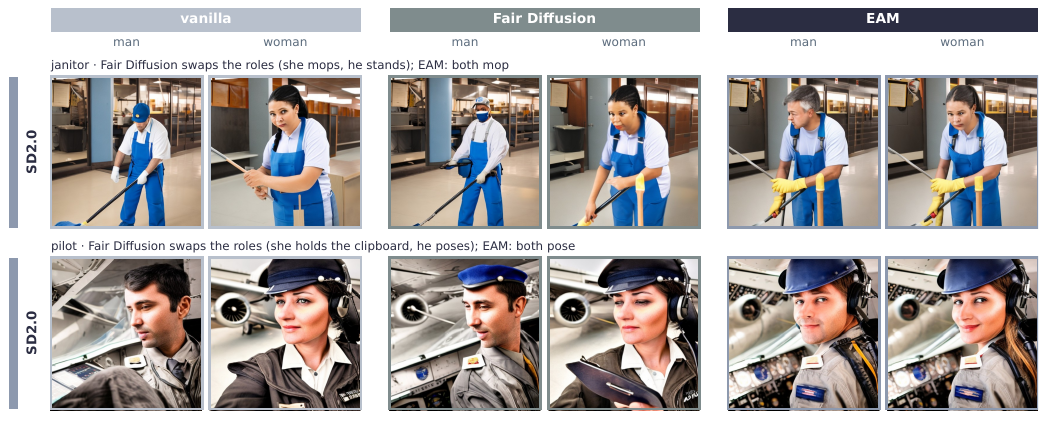}{\linewidth}
\caption{\textbf{Fair Diffusion and \eam on the same seeds} (SD2.0, janitor and
pilot), each a man and woman pair next to vanilla.}
\label{fig:s-baselines}
\end{figure*}

\paragraph{Fair Diffusion on the same seeds.}
On SD2.0, Fair Diffusion swaps the roles instead of sharing them: the woman mops
while the man stands (janitor), and she holds the clipboard while he poses
(pilot). \eam gives both the mop in the first pair, but in the second both
pose, an instance of the levelling-down of \cref{fig:s-engage}
(\cref{fig:s-baselines}).

\paragraph{Picking versus averaging during sampling.}
When one gender's prompt drives the first $K$ steps, its layout and often the
person carry into the final image; when \eam averages the two prompts over the
same steps, the layout is shared and the requested person appears
(\cref{fig:s-traj,fig:s-switchcurve}).

\begin{figure*}[h]
\centering
\suppimg{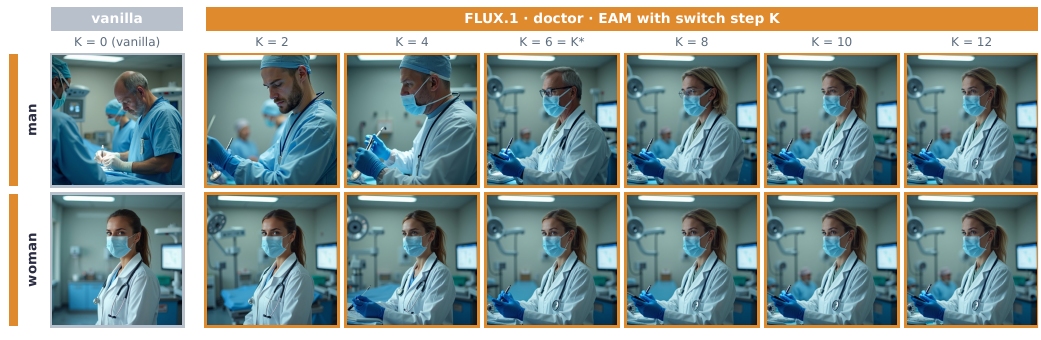}{\linewidth}
\caption{\textbf{One pair across switch steps} on FLUX.1 (doctor), from $K = 0$
(vanilla) to $K = 12$, twice $K^\star$.}
\label{fig:s-kvis}
\end{figure*}

\paragraph{Effect of the switch step.}
As $K$ grows the woman takes over the working layout within the first steps, and
on FLUX.1 past $K^\star$ the requested gender starts to be lost
(\cref{fig:s-kvis}), as \cref{fig:s-ksweep} predicts.

\begin{figure*}[t]
\centering
\suppimg{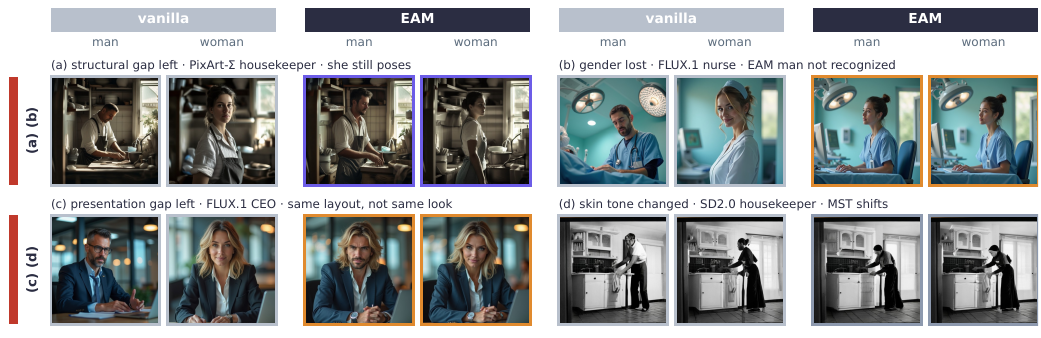}{\linewidth}
\caption{\textbf{Where \eam falls short.} (a)~Remaining structural gap. (b)~Lost
gender. (c)~Presentation gap on a shared layout. (d)~Changed skin tone.}
\label{fig:s-failures}
\end{figure*}

\paragraph{Failure cases and what remains.}
\Cref{fig:s-failures} shows a structural gap that \eam leaves (PixArt-$\Sigma$
housekeeper), a pair in which the requested man is no longer recognized (FLUX.1
nurse), a shared layout that still differs in expression and attire (FLUX.1 CEO),
and a change of perceived skin tone (SD2.0 housekeeper).

\section{Limitations and broader impact}
\label{sec:s-limits}

\paragraph{Scope.}
Binary gender as requested, 20 U.S.-defined occupations, one template, one
person per image, and four open generators. Non-binary people, other roles and
languages, multi-person scenes, and commercial generators are not covered.

\paragraph{Measurement.}
The vocabulary, closed answers, and fidelities depend on models with their own
biases. Gender-blind instructions, instruments, and a second describer reduce
this dependence but do not remove it, and no human annotation was used. Gender
fidelity is the gender a model perceives, not the gender of any person.

\paragraph{What \eam changes.}
\eam keeps the requested gender but softens its marking, leaving more women
visually ambiguous to a second describer (21\% vs.\ 13\%), changes perceived skin
tone in about half of the pairs, and adds jewelry to women slightly more often.
It closes the engagement gap partly by drawing men at work less often, leaves
most of the presentation gap, and does not address who appears when the prompt
leaves identity open.

\paragraph{Broader impact.}
\pose makes a less visible bias measurable, and \eam reduces it at inference
time without retraining. The same measurements could amplify a stereotype by
following one gender's early steps instead of the average.

\end{document}